\documentclass[11pt, a4paper, logo, copyright]{eai}
\usepackage{shlab}

\usepackage[authoryear, sort&compress, round]{natbib}
\usepackage[]{mdframed}

\usepackage{dblfloatfix}

\usepackage{amsmath,amsfonts,bm}
\usepackage{multirow}
\usepackage{subcaption}
\usepackage{wrapfig}

\def\eqref#1{equation~\ref{#1}}

\def\1{\bm{1}}

\DeclareMathAlphabet{\mathsfit}{\encodingdefault}{\sfdefault}{m}{sl}
\SetMathAlphabet{\mathsfit}{bold}{\encodingdefault}{\sfdefault}{bx}{n}

\usepackage{listings}
\usepackage{tikz}
\usetikzlibrary{arrows.meta}
\usepackage{hyperref}
\usepackage{url}
\usepackage{graphicx}
\usepackage{amsmath} %
\usepackage{mathrsfs} %
\usepackage{etoolbox}
\usepackage{cleveref}
\usepackage{tcolorbox}
\usepackage{colortbl}
\usepackage{booktabs}       %
\usepackage{amsfonts}       %
\usepackage{amssymb}
\usepackage{nicefrac}       %
\usepackage{microtype}      %
\usepackage{caption}
\usepackage{xspace} %
\usepackage{subcaption}
\usepackage{algorithm}
\usepackage{algorithmic}
\usepackage{multirow}
\usepackage{subcaption}
\usepackage{lipsum}

\usepackage{adjustbox}

\usepackage{booktabs} %
\usepackage{enumitem}%
\setlist[itemize]{noitemsep, topsep=0pt}
\usepackage{enumitem,kantlipsum}

\usepackage{multicol,multirow} 
\usepackage{graphicx}    
\usepackage{pifont}
\usepackage{minted}
\usepackage{ulem}
\usepackage{xcolor}
\usepackage{longtable}
\usepackage{makecell} 
\usepackage{listings}
\usepackage{xcolor}
\usepackage[T1]{fontenc}

\lstdefinestyle{yamlstyle}{
    basicstyle=\ttfamily\footnotesize,
    numbers=left,
    numberstyle=\tiny,
    stepnumber=1,
    numbersep=6pt,
    frame=single,
    framerule=0.5pt,
    breaklines=true,
    breakatwhitespace=true,
    tabsize=2,
    captionpos=b,
    keywordstyle=\color{blue},
    commentstyle=\color{gray},
    stringstyle=\color{teal},
}

\newlength\savewidth

\definecolor{baselinecolor}{HTML}{d6eaf8}

\definecolor{mygray}{gray}{0.4}

\AtBeginEnvironment{tcolorbox}{\tiny}

\newcount\Comments  %
\usepackage{color}
\definecolor{darkred}{rgb}{0.9,0,0}
\definecolor{darkgreen}{rgb}{0,0.5,0}
\definecolor{darkblue}{rgb}{0,0,0.7}
\definecolor{purple}{rgb}{.6, 0,.6}
\definecolor{orange}{rgb}{1.0,0.64,0}

\definecolor{deemph}{gray}{0.6}

\definecolor{baselinecolor}{gray}{.9}
\definecolor{yellow}{RGB}{218,165,32}
\definecolor{lightcyan}{rgb}{0.88, 1.0, 1.0}
\definecolor{lightskyblue}{rgb}{0.53, 0.81, 0.98}
\definecolor{aliceblue}{rgb}{0.94, 0.97, 1.0}
\definecolor{LightSlateBlue}{RGB}{70,130,180}
\definecolor{DeepBlue}{RGB}{65,100,170}
\definecolor{DeepPurple}{RGB}{136,105,160}
\definecolor{LightGreen}{RGB}{59,125,35}
\definecolor{LightRed}{RGB}{234,66,53}
\definecolor{cvprblue}{rgb}{0.21,0.49,0.74}

\definecolor{mypink}{RGB}{254,102,140}
\definecolor{myclay}{RGB}{59,182,176}

\usepackage{pgfplots}
\pgfplotsset{compat=1.18}

\newcommand{\kibitz}[2]{\ifnum\Comments=1\textcolor{#1}{#2}\fi}

\definecolor{bgfirst}{RGB}{255, 209, 92}
\definecolor{bgsecond}{RGB}{255, 226, 150}
\definecolor{bgthird}{RGB}{255, 242, 207}
\newcommand{\first}[1]{\cellcolor{bgfirst}\textbf{#1}}
\newcommand{\second}[1]{\cellcolor{bgsecond}#1}
\newcommand{\third}[1]{\cellcolor{bgthird}#1}

\title{GameRealm: Grounding Visual Learning in Interactive Games}

\renewcommand{\today}{}

\DeclareRobustCommand{\GameRealmAuthors}{%
  \begingroup
  \normalfont\small
  \begin{tabular}{@{}c@{}}
    Weiguang Zhao\textsuperscript{1}\enspace
    Yifan Wang\textsuperscript{1}\enspace
    Ni Yang\textsuperscript{1}\enspace
    Yating Wang\textsuperscript{1}\enspace
    Xingyu Miao\textsuperscript{1}\enspace
    Yankai Jiang\textsuperscript{1}
    \tabularnewline[3pt]
    Mulin Yu\textsuperscript{1}\enspace
    Tao Lu\textsuperscript{1}\enspace
    Haoyu Guo\textsuperscript{1*}\enspace
    Junting Dong\textsuperscript{1*}\enspace
    Weinan Zhang\textsuperscript{1,2}\enspace
    Chunhua Shen\textsuperscript{1,3}
    \tabularnewline[6pt]
    \textsuperscript{1}Shanghai Artificial Intelligence Laboratory\quad
    \textsuperscript{2}Shanghai Jiao Tong University\quad
    \textsuperscript{3}Zhejiang University
    \tabularnewline[4pt]
    {\footnotesize\textbf{Project page:}~
     \href{weiguangzhao.github.io/GameRealm}{weiguangzhao.github.io/GameRealm}}
  \end{tabular}%
  \endgroup
}

\author{\GameRealmAuthors}

\renewcommand{\footnoterule}{%
  \kern -3pt
  \hrule width 1.0\columnwidth height 0.5pt
  \kern 2.6pt
}

\usepackage{placeins}

\begin{document}

\begin{abstract}
Visual foundation models are central to physical intelligence, yet their development remains constrained by fragmented training datasets, inconsistent annotation quality, and limited multimodal supervision. To address these challenges, we introduce GameRealm, a large-scale, richly annotated dataset that grounds visual observations in synchronized, game-native supervision. GameRealm encompasses over 120 million RGB frames and 1,000 hours of gameplay footage across all views. Beyond its unprecedented scale, GameRealm provides rich multimodal supervision, including metric depth, camera parameters, synchronized audio, control inputs, and 3D point trajectories. Its diversity further extends to flexible view configurations, spanning single-view, stereo, multi-view, and panoramic recordings. To enable scalable acquisition across diverse game environments, we develop a unified data collection pipeline that combines human-operated interaction with agentic exploration. Extensive experiments across diverse visual learning tasks demonstrate that GameRealm consistently strengthens visual foundation models in both visual generation and geometric learning. These results establish GameRealm as a versatile and valuable data resource, unlocking new opportunities for advancing visual intelligence.

\end{abstract}

\maketitle

\section{Introduction}

The pursuit of physical intelligence requires visual foundation models to understand not only the appearance of the world, but also its underlying geometry, dynamics, and interactions. Developing such capabilities demands large-scale, diverse, and richly supervised visual experiences. Modern video games offer a compelling source of such data, combining visually rich environments, complex scene dynamics, and interactive behaviors. More importantly, game engines provide direct access to precise, game-native supervision, including metric depth, camera poses, and motion information~\citep{li2026sekai}, which remains difficult to acquire accurately and at scale in the real world. This unique combination of visual diversity, interactive dynamics, and geometric precision positions game environments as a powerful foundation for advancing visual intelligence.

Despite growing demands for large-scale visual data, existing datasets remain fragmented across tasks and supervision modalities. Interactive world models require temporally aligned visual observations, control inputs, and audio to learn action-conditioned dynamics, whereas geometric foundation models depend on accurate depth, camera poses, and stereo or multi-view observations for 3D/4D reconstruction~\citep{wang2025vggt,zhang2026efficiently}. However, as summarized in Tab.~\ref{tab:dataset_comparison}, existing datasets rarely satisfy these requirements simultaneously. Conventional synthetic datasets~\citep{wang2020tartanair,roberts2021hypersim,zheng2023pointodyssey}, including SynthVerse~\citep{Zhao2026SynthVerse}, often depend on manually constructed environments, constraining visual diversity and scalability. Recent gameplay datasets offer richer environments but remain limited in supervision. SeKai-Game~\citep{li2026sekai} lacks depth annotations, while OmniWorld-Game~\citep{zhou2025omniworld} relies on $\pi^3$-estimated~\citep{wang2026pi} camera trajectories rather than  ground truth. Neither provides synchronized multi-view observations or audio, highlighting the need for a unified, scalable dataset with diverse environments, rich multimodal supervision, and flexible view configurations to advance both visual generation and geometric learning.

To address these limitations, we introduce GameRealm, a large-scale game-derived dataset that integrates geometric ground truth, synchronized multimodal signals, and diverse visual observations across single-view, stereo, multi-view, and panoramic configurations. GameRealm comprises over 120 million frames and 1,000 hours of gameplay across diverse environments and interaction scenarios. We combine human-operated collection to capture diverse and context-dependent interactions with automated exploration to systematically navigate free-roaming environments. Through dedicated game-engine plugins, our acquisition pipeline directly accesses ground truth, establishing precise correspondence between visual observations and their underlying geometric and control information. The collected data include synchronized RGB images, metric depth, camera parameters, audio,  keyboard and mouse inputs, and point tracking. Furthermore, we filter visually degraded and inconsistent clips.

On the other hand, we examine the utility of GameRealm through multiple visual learning tasks. Employing the recorded player inputs, we fine-tune the world model to predict how the player’s view evolves under a sequence of keyboard and mouse commands. We then study the geometric information in these recordings by fine-tuning models for stereo reconstruction and multi-view reconstruction. Evaluation on public real-world benchmarks tests how well the geometric supervision acquired in games transfers beyond the environments in which it was collected. Our contributions are as follows:

\begin{figure*}[t]
\vspace{-1.0em}
  \includegraphics[width=0.99\textwidth]{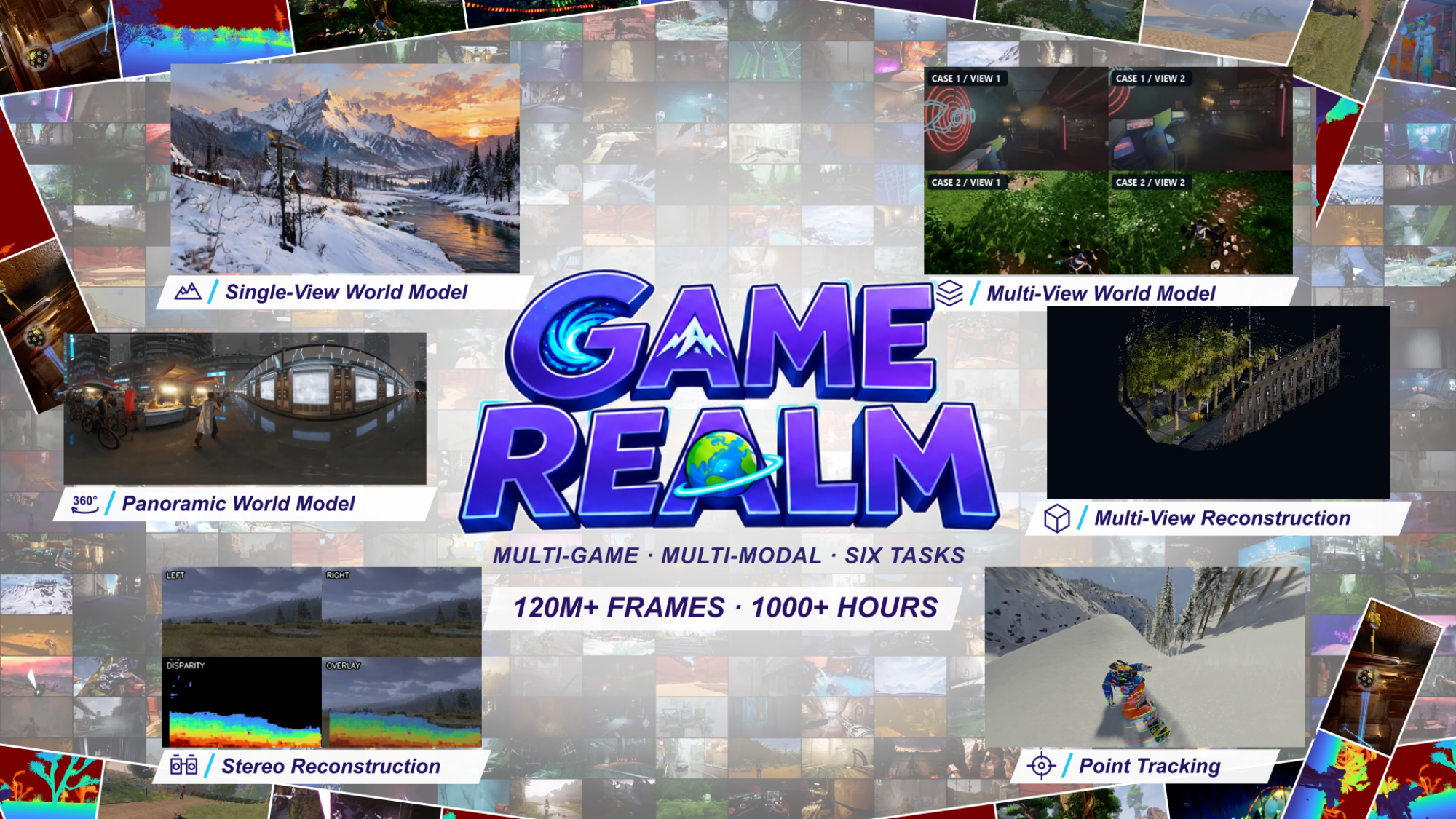}
  \caption{GameRealm overview. GameRealm spans diverse games and supports multiple tasks: single-view, multi-view, and panoramic world modeling, alongside stereo, multi-view reconstruction, and point tracking. The collage showcases original game ground-truth data totaling over 120 million frames and 1,000 hours.}
  \label{fig:pipeline}
  \vspace{-1.0em}
\end{figure*}

\begin{itemize}
    \item We introduce GameRealm, a large-scale game-derived dataset featuring 110M+ frames, 1,000+ hours of gameplay, synchronized multimodal ground truth, and diverse viewing configurations.
    \item We develop an game-grounded acquisition pipeline, integrating human interaction, agentic exploration, and multi-view-aware quality assessment for scalable and reliable data collection.
    \item We validate GameRealm across diverse downstream tasks, spanning interactive world modeling and geometric reconstruction, demonstrating its versatility and cross-domain generalization.
    \item We propose a point tracking benchmark for 2D and 3D point tracking across diverse motions, targets, and environments, with eight subcategories grouped under Interaction and Locomotion.
\end{itemize}

\section{Related Work}

\subsection{Datasets}

Synthetic datasets provide geometric and motion supervision at scale. MPI Sintel~\citep{butler2012naturalistic}, Spring~\citep{mehl2023spring}, and Dynamic Replica~\citep{karaev2023dynamicreplica} support optical flow, stereo, and dynamic scene understanding, while TartanAir~\citep{wang2020tartanair}, BlendedMVS~\citep{yao2020blendedmvs}, Hypersim~\citep{roberts2021hypersim}, and EDEN~\citep{le2021eden} target scene geometry and navigation. For temporal correspondence, FlyingThings++~\citep{harley2022particle} and PointOdyssey~\citep{zheng2023pointodyssey} provide point-tracking supervision. Recent datasets, including Syn4D~\citep{jiang2026syn4d} and SynthVerse~\citep{Zhao2026SynthVerse}, further extend synthetic data toward synchronized multi-view observations, dense 4D correspondence, and diverse interactions.

\begin{table*}[ht]
    \centering
    \small
    \setlength{\tabcolsep}{5pt}
    \renewcommand{\arraystretch}{1.08}
    \resizebox{\textwidth}{!}{%
    \begin{tabular}{l r c c c c c c c c c c}
        \toprule
        Dataset & \# Frames
        & \multicolumn{6}{c}{Data modality}
        & \multicolumn{4}{c}{View} \\
        \cmidrule(lr){3-8}
        \cmidrule(lr){9-12}
        & & Depth & Camera & Text & Audio & Keypress & Tracking
        & Single-view & Stereo & Multi-view & Panoramic \\
        \midrule
        MPI Sintel
        & 1K
        & \textcolor{green!60!black}{\ding{52}}
        & \textcolor{green!60!black}{\ding{52}}
        & \textcolor{red}{\ding{56}}
        & \textcolor{red}{\ding{56}}
        & \textcolor{red}{\ding{56}}
        & \textcolor{red}{\ding{56}}
        & \textcolor{green!60!black}{\ding{52}}
        & \textcolor{green!60!black}{\ding{52}}
        & \textcolor{red}{\ding{56}}
        & \textcolor{red}{\ding{56}} \\

        FlyingThings++
        & 28K
        & \textcolor{green!60!black}{\ding{52}}
        & \textcolor{red}{\ding{56}}
        & \textcolor{red}{\ding{56}}
        & \textcolor{red}{\ding{56}}
        & \textcolor{red}{\ding{56}}
        & \textcolor{green!60!black}{\ding{52}}
        & \textcolor{green!60!black}{\ding{52}}
        & \textcolor{green!60!black}{\ding{52}}
        & \textcolor{red}{\ding{56}}
        & \textcolor{red}{\ding{56}} \\

        TartanAir
        & 1,000K
        & \textcolor{green!60!black}{\ding{52}}
        & \textcolor{green!60!black}{\ding{52}}
        & \textcolor{red}{\ding{56}}
        & \textcolor{red}{\ding{56}}
        & \textcolor{red}{\ding{56}}
        & \textcolor{red}{\ding{56}}
        & \textcolor{green!60!black}{\ding{52}}
        & \textcolor{green!60!black}{\ding{52}}
        & \textcolor{red}{\ding{56}}
        & \textcolor{green!60!black}{\ding{52}} \\

        BlendedMVS
        & 17K
        & \textcolor{green!60!black}{\ding{52}}
        & \textcolor{green!60!black}{\ding{52}}
        & \textcolor{red}{\ding{56}}
        & \textcolor{red}{\ding{56}}
        & \textcolor{red}{\ding{56}}
        & \textcolor{red}{\ding{56}}
        & \textcolor{red}{\ding{56}}
        & \textcolor{red}{\ding{56}}
        & \textcolor{green!60!black}{\ding{52}}
        & \textcolor{red}{\ding{56}} \\

        HyperSim
        & 77K
        & \textcolor{green!60!black}{\ding{52}}
        & \textcolor{green!60!black}{\ding{52}}
        & \textcolor{red}{\ding{56}}
        & \textcolor{red}{\ding{56}}
        & \textcolor{red}{\ding{56}}
        & \textcolor{red}{\ding{56}}
        & \textcolor{green!60!black}{\ding{52}}
        & \textcolor{red}{\ding{56}}
        & \textcolor{red}{\ding{56}}
        & \textcolor{red}{\ding{56}} \\

        Dynamic Replica
        & 169K
        & \textcolor{green!60!black}{\ding{52}}
        & \textcolor{green!60!black}{\ding{52}}
        & \textcolor{red}{\ding{56}}
        & \textcolor{red}{\ding{56}}
        & \textcolor{red}{\ding{56}}
        & \textcolor{green!60!black}{\ding{52}}
        & \textcolor{red}{\ding{56}}
        & \textcolor{green!60!black}{\ding{52}}
        & \textcolor{red}{\ding{56}}
        & \textcolor{red}{\ding{56}} \\

        Spring
        & 23K
        & \textcolor{green!60!black}{\ding{52}}
        & \textcolor{green!60!black}{\ding{52}}
        & \textcolor{red}{\ding{56}}
        & \textcolor{red}{\ding{56}}
        & \textcolor{red}{\ding{56}}
        & \textcolor{red}{\ding{56}}
        & \textcolor{red}{\ding{56}}
        & \textcolor{green!60!black}{\ding{52}}
        & \textcolor{red}{\ding{56}}
        & \textcolor{red}{\ding{56}} \\

        EDEN
        & 300K
        & \textcolor{green!60!black}{\ding{52}}
        & \textcolor{green!60!black}{\ding{52}}
        & \textcolor{red}{\ding{56}}
        & \textcolor{red}{\ding{56}}
        & \textcolor{red}{\ding{56}}
        & \textcolor{red}{\ding{56}}
        & \textcolor{green!60!black}{\ding{52}}
        & \textcolor{red}{\ding{56}}
        & \textcolor{red}{\ding{56}}
        & \textcolor{red}{\ding{56}} \\

        PointOdyssey
        & 216K
        & \textcolor{green!60!black}{\ding{52}}
        & \textcolor{green!60!black}{\ding{52}}
        & \textcolor{red}{\ding{56}}
        & \textcolor{red}{\ding{56}}
        & \textcolor{red}{\ding{56}}
        & \textcolor{green!60!black}{\ding{52}}
        & \textcolor{green!60!black}{\ding{52}}
        & \textcolor{red}{\ding{56}}
        & \textcolor{green!60!black}{\ding{52}}
        & \textcolor{red}{\ding{56}} \\

        Syn4D
        & 1.4M
        & \textcolor{green!60!black}{\ding{52}}
        & \textcolor{green!60!black}{\ding{52}}
        & \textcolor{red}{\ding{56}}
        & \textcolor{red}{\ding{56}}
        & \textcolor{red}{\ding{56}}
        & \textcolor{green!60!black}{\ding{52}}
        & \textcolor{green!60!black}{\ding{52}}
        & \textcolor{red}{\ding{56}}
        & \textcolor{green!60!black}{\ding{52}}
        & \textcolor{red}{\ding{56}}\\

        SynthVerse
        & 5.8M
        & \textcolor{green!60!black}{\ding{52}}
        & \textcolor{green!60!black}{\ding{52}}
        & \textcolor{red}{\ding{56}}
        & \textcolor{red}{\ding{56}}
        & \textcolor{red}{\ding{56}}
        & \textcolor{green!60!black}{\ding{52}}
        & \textcolor{green!60!black}{\ding{52}}
        & \textcolor{red}{\ding{56}}
        & \textcolor{green!60!black}{\ding{52}}
        & \textcolor{red}{\ding{56}}\\

        SeKai-Game
        & 4.3M
        & \textcolor{red}{\ding{56}}
        & \textcolor{green!60!black}{\ding{52}}
        & \textcolor{green!60!black}{\ding{52}}
        & \textcolor{red}{\ding{56}}
        & \textcolor{red}{\ding{56}}
        & \textcolor{red}{\ding{56}}
        & \textcolor{green!60!black}{\ding{52}}
        & \textcolor{red}{\ding{56}} 
        & \textcolor{red}{\ding{56}}
        & \textcolor{red}{\ding{56}}\\

        OmniWorld-Game
        & 18.5M
        & \textcolor{green!60!black}{\ding{52}}
        & \textcolor{green!60!black}{\ding{52}}
        & \textcolor{green!60!black}{\ding{52}}
        & \textcolor{red}{\ding{56}}
        & \textcolor{red}{\ding{56}}
        & \textcolor{red}{\ding{56}}
        & \textcolor{green!60!black}{\ding{52}}
        & \textcolor{red}{\ding{56}}
        & \textcolor{red}{\ding{56}}
        & \textcolor{red}{\ding{56}}\\

        \textbf{GameRealm (Ours)}
        & 120M+
        & \textcolor{green!60!black}{\ding{52}}
        & \textcolor{green!60!black}{\ding{52}}
        & \textcolor{green!60!black}{\ding{52}}
        & \textcolor{green!60!black}{\ding{52}}
        & \textcolor{green!60!black}{\ding{52}}
        & \textcolor{green!60!black}{\ding{52}}
        & \textcolor{green!60!black}{\ding{52}}
        & \textcolor{green!60!black}{\ding{52}}
        & \textcolor{green!60!black}{\ding{52}}
        & \textcolor{green!60!black}{\ding{52}}\\
        \bottomrule
    \end{tabular}%
    }
\caption{Comparison between Gamerealm and existing synthetic datasets. }
\label{tab:dataset_comparison}
\end{table*}

Game-derived datasets offer a complementary source of realistic visual observations and interactive supervision. Sekai-Game~\citep{li2026sekai} provides gameplay videos with engine-derived camera trajectories, while OmniWorld-Game~\citep{zhou2025omniworld} combines renderer-derived depth with estimated camera poses and generated annotations. Despite these advances, existing resources provide limited combinations of multi-view observations, audio, and recorded control inputs. Moreover, estimated annotations and camera trajectories do not provide the same supervision as game-native ground truth and actual player actions. As summarized in Table~\ref{tab:dataset_comparison}, GameRealm brings these capabilities together through synchronized RGB, metric depth, camera parameters, control inputs, audio, and point trajectories across single-view, stereo, multi-view, and panoramic settings. Its acquisition pipeline combines human-operated interaction capture with automated exploration, providing a unified resource for interactive world modeling and visual geometry learning.

\subsection{Interactive World Model}

Interactive world models extend video generation to action-conditioned prediction of dynamic environments. Genie~\citep{bruce2024genie} learns latent actions from unlabeled videos, while DIAMOND~\citep{alonso2024diffusion} and GameNGen~\citep{valevski2025diffusion} demonstrate diffusion-based and autoregressive approaches to interactive visual simulation. More recent methods, including Matrix-Game~\citep{zhang2025matrix} and Matrix-Game 2.0~\citep{he2025matrix}, introduce explicit keyboard and mouse conditioning for controllable, real-time gameplay generation. These works establish action--observation alignment as a key requirement for interactive world modeling.

Beyond short-term action following, recent methods focus on long-horizon consistency and persistent spatial memory. WorldPlay~\citep{sun2025worldplay} employs reconstructed context memory, while LingBot-World 2.0~\citep{team2026advancing,gao2026infinite}, AlayaWorld~\citep{team2026alayaworld}, and ReWorld~\citep{chen2026reworld} explore causal generation, compressed history, and geometry-aware memory for extended interaction. Meanwhile, panoramic world models such as PanoWorld~\citep{li2026panoworld} address spatial continuity under omnidirectional viewing, and LTX-2~\citep{hacohen2026ltx} jointly generates synchronized video and audio. Together, these advances extend world modeling toward controllable, spatially consistent, and multimodal generation. GameRealm complements these model-centric advances with synchronized, engine-grounded supervision. It records visual observations alongside available player inputs, camera parameters, depth, and audio, enabling the adaptation of pretrained models such as Helios~\citep{yuan2026helios} and LTX-2.3 to action-conditioned perspective and panoramic generation. This explicitly connects interactive generation with geometric and multimodal information.

\subsection{Visual Geometry Grounded Learning}

Visual geometry learning aims to recover scene structure and motion from visual observations~\citep{wang2025vggt,wang2026vggt,miao2026trajvg}. We focus on three complementary tasks: stereo reconstruction, multi-view reconstruction, and point tracking, which respectively recover depth, spatial structure, and persistent temporal correspondences.

\subsubsection{Stereo Reconstruction}

Stereo reconstruction estimates dense disparity from calibrated image pairs and converts it into metric depth. Early learning-based methods, such as RAFT-Stereo~\citep{lipson2021raft} and IGEV-Stereo~\citep{xu2023iterative}, improve correspondence estimation through recurrent refinement and explicit geometric encoding. Recent approaches further incorporate monocular priors to handle ambiguous regions: Stereo Anywhere~\citep{bartolomei2025stereo}, MonSter~\citep{cheng2025monster}, and DEFOM-Stereo~\citep{jiang2025defom} combine stereo matching with pretrained visual or depth representations. FoundationStereo~\citep{wen2025foundationstereo} and Fast-FoundationStereo~\citep{wen2026fast} extend this direction toward robust zero-shot generalization and real-time inference.

\subsubsection{Multi-view Reconstruction}

Traditional structure-from-motion and multi-view stereo methods~\citep{schonberger2016structure,schonberger2016pixelwise} recover camera poses and geometry through feature matching and geometric optimization. Recent feed-forward models simplify this process: DUSt3R~\citep{wang2024dust3r} and MASt3R~\citep{leroy2024grounding} establish pairwise geometric reconstruction, while Fast3R~\citep{yang2025fast3r}, CUT3R~\citep{wang2025continuous}, and VGGT~\citep{wang2025vggt,wang2026vggt} extend reconstruction to multiple views and online perception. More recent models, including $\pi^3$~\citep{wang2026pi}, MapAnything~\citep{keetha2026mapanything}, and Depth Anything 3~\citep{lin2025depth}, further improve flexibility across input configurations and geometric representations. Nevertheless, robustness to dynamic scenes, complex motion, and domain shifts remains challenging. GameRealm provides synchronized multi-view observations, metric depth, and per-frame camera parameters for geometric learning. We investigate their value by fine-tuning a pretrained $\pi^3$ model while retaining its original objectives and evaluation protocol, isolating the contribution of game-derived supervision to cross-domain reconstruction.

\subsubsection{Point Tracking}

Point tracking establishes persistent correspondences across video frames under camera motion, occlusion, and object deformation. 2D methods such as TAPIR~\citep{doersch2023tapir} and CoTracker3~\citep{karaev2025cotracker3} improve point localization and temporal consistency, while recent 3D trackers explicitly model scene geometry and camera motion. SpatialTrackerV2~\citep{xiao2025spatialtrackerv2} jointly estimates 3D trajectories and camera motion, whereas TAPIP3D~\citep{zhang2026tapip3d} tracks points in persistent 3D space. Correspondingly, TAPVid-3D~\citep{koppula2024tapvid} and SynthVerse~\citep{Zhao2026SynthVerse} provide real-world and synthetic benchmarks for 3D tracking. Unlike projected 2D trajectories, world-space point positions provide direct supervision for separating object motion from camera motion. GameRealm records synchronized 3D point positions, 2D projections, depth, and camera parameters where supported by the acquisition pipeline, enabling the study of point tracking under diverse interactions and viewpoint changes.

\section{GameRealm Dataset}
\subsection{Data Acquisition}
\label{sec:acquisition}
\begin{figure*}[ht]
  \includegraphics[width=0.99\textwidth]{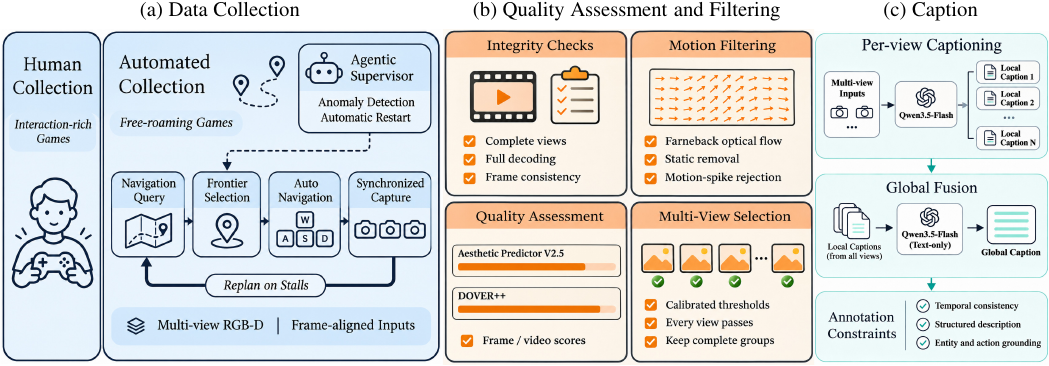}
  \caption{Data acquisition pipeline}
  \label{fig:Data_Acquisition}
\end{figure*}

 We develop a unified game-data collection pipeline that integrates human-driven and automated data acquisition, as illustrated in Fig.~\ref{fig:Data_Acquisition}. For interaction-rich scenarios, human operators provide direct control to capture complex interactions, while free-roaming scenarios are handled by an automated navigation system with frontier-based exploration and adaptive replanning. An agentic supervisor further monitors the automated process, detecting anomalies and triggering restarts when necessary. Data collected through both pathways undergo unified quality assessment and filtering to ensure consistency and reliability. The resulting dataset provides synchronized, game-native multimodal ground truth, including multi-view RGB images, metric depth, camera parameters, audio, keyboard and mouse inputs, and point tracking, supporting diverse downstream spatial understanding tasks.

\subsubsection{Data Collection}
\begin{figure*}[ht]
  \includegraphics[width=0.9775\textwidth]{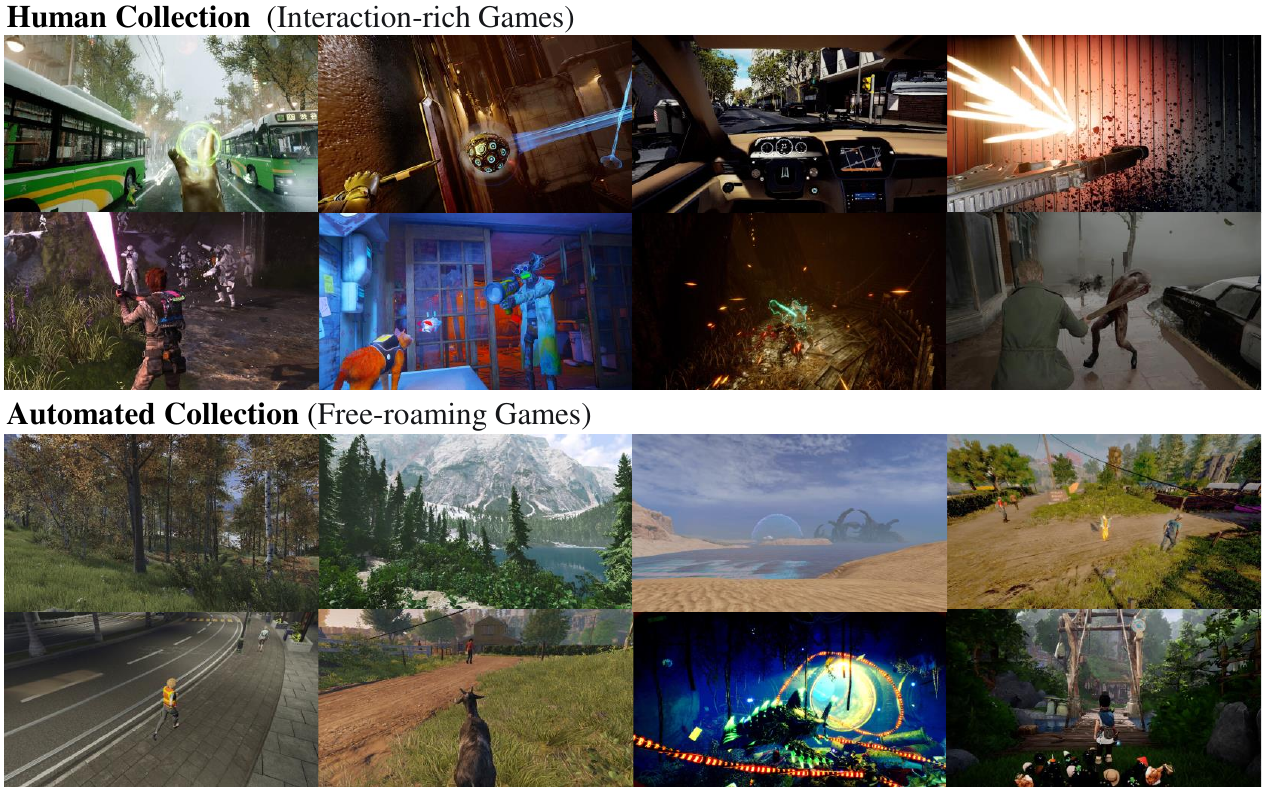}
  \caption{Data collection. GameRealm encompasses diverse game environments, first- and third-person perspectives, and a wide range of playable characters, including humans and animals. Data are acquired through two complementary strategies: human-driven collection for interaction-rich scenarios (top) and automated collection for free-roaming scenarios (bottom)}
  \label{fig:Data_Collection}
\end{figure*}

We implement a dedicated game-engine plugin that provides direct access to engine-level ground-truth (GT) information, supporting two complementary acquisition paradigms, as shown in Fig.~\ref{fig:Data_Collection}~(a): human-driven collection for gameplay involving complex, context-sensitive interactions, and automated collection for free-roaming scenarios that benefit from systematic navigation and exploration.

\noindent\textbf{Ground-Truth Acquisition.} The plugin directly captures game-native metric depth and camera parameters, following established practices in datasets such as TartanAir~\citep{wang2020tartanair}, and SeKai-Game~\citep{li2026sekai}. Game-engine timestamps ensure temporal synchronization across modalities and viewpoints. We evaluate cross-view alignment on view pairs randomly sampled from GameRealm clips and report the reprojection errors in Tab.~\ref{tab:cross_view_alignment}.

\noindent\textbf{Human Collection.} Human-operated collection primarily targets games with rich, context-dependent interactions. Representative activities include performing extreme-sports maneuvers, aiming and firing weapons, traversing complex terrain, and picking up or manipulating objects. These activities often require precise timing, coordinated controls, and context-aware decisions. Human operators execute meaningful gameplay sequences while the acquisition system records the corresponding observations and keyboard and mouse inputs. The resulting recordings follow the same unified ground-truth representation as automated collection, enabling both sources to be integrated into a common dataset.

\noindent\textbf{Automated Collection.} Automated collection primarily targets free-roaming games, where navigation provides a practical means of acquiring diverse scene observations. The collection workflow consists of navigation queries, frontier selection, automatic navigation, and synchronized capture. Game-native navigation queries identify reachable candidate destinations, and frontier selection prioritizes endpoints that extend spatial exploration. A closed-loop controller then updates keyboard and mouse inputs according to the player's position and route progress. Movement stalls trigger route replanning to support continued exploration.

During navigation, the acquisition system records multi-view RGB-D  together with frame-aligned inputs. Timestamps are recorded for each captured observation and its corresponding inputs, enabling temporal synchronization across RGB images, depth maps, camera parameters, and control signals. An agentic supervisor is integrated into the automated collection workflow. It monitors runtime status, detects operational anomalies, and initiates restart procedures when needed. Its role is to manage collection failures, while navigation queries and the feedback controller govern gameplay movement.

\subsection{Data Quality Assessment}
\label{sec:data_quality_filtering}

\begin{figure*}[b!]
    \centering
    \includegraphics[width=0.9775\textwidth]{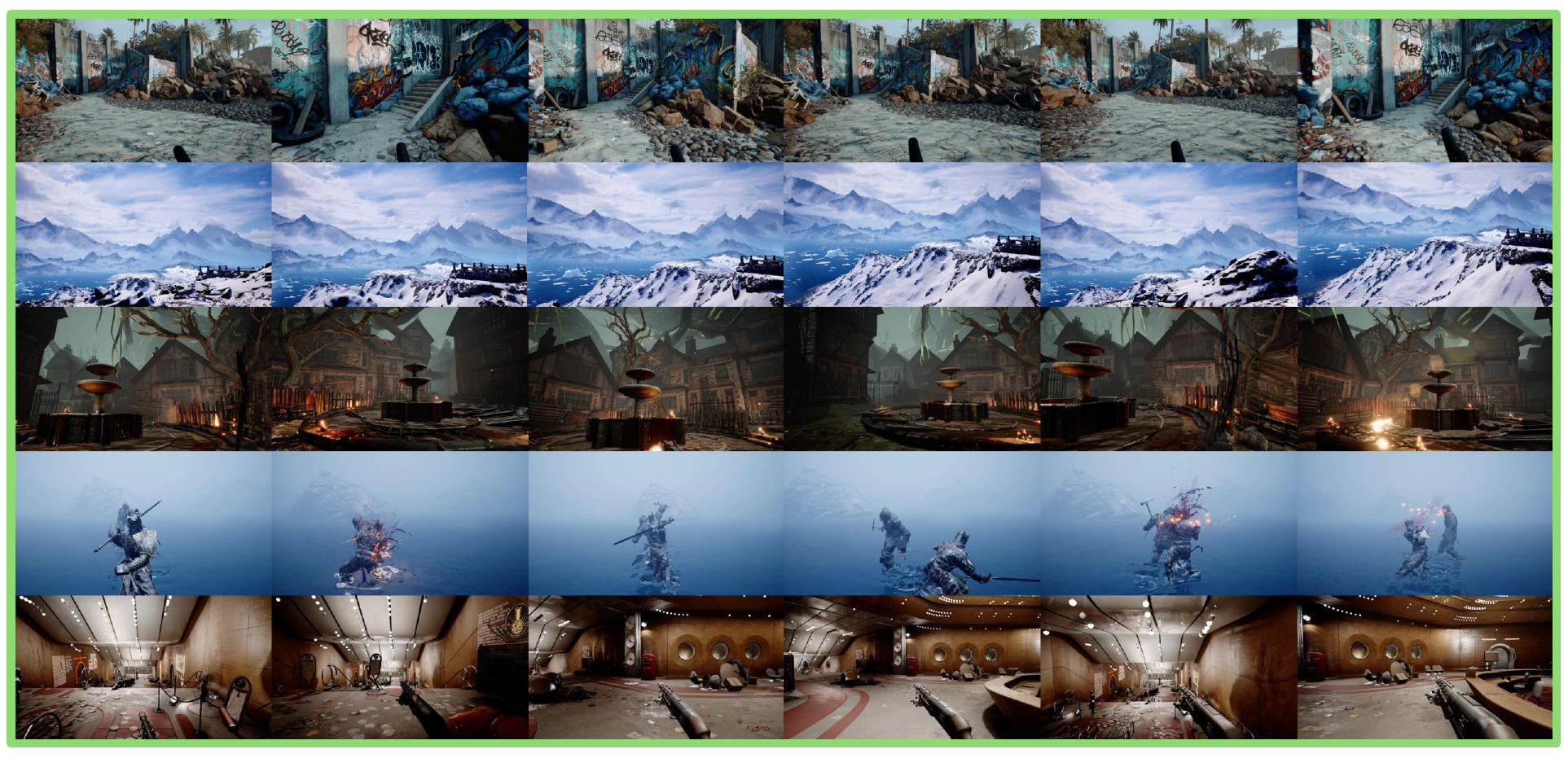}
    \caption{
    Examples of retained video clips after quality filtering.
    The selected clips maintain clear scene visibility and coherent temporal progression across diverse game environments, providing reliable visual observations for downstream spatial understanding tasks.
    }
    \label{fig:data_true}
\end{figure*}

We first validate video integrity and motion consistency, as illustrated in Fig.~\ref{fig:Data_Collection}~(b). All expected views must be present, and each video is fully decoded to verify its frame count, spatial dimensions, and frame rate against the corresponding annotations and acquisition settings. Groups with missing views or failed integrity checks are excluded. We then resize video frames to a unified resolution and sample them at fixed intervals to estimate dense Farneb\"ack optical flow. For each sampled frame pair, we compute the spatially averaged flow magnitude and normalize it by the shorter image dimension. Clips with insufficient temporal motion or an excessively high ratio of maximum to mean motion magnitude, potentially indicating residual scene cuts or transitions, are excluded.

We assess visual quality using complementary image-based and video-based predictors. Aesthetic Predictor V2.5, built on a SigLIP encoder, scores frames sampled at one frame per second; their mean provides the aesthetic score $A_v$ for view $v$. DOVER++~\citep{wu2023exploring}, in turn, evaluates the video through technical and aesthetic quality branches. The branch outputs are calibrated, combined, and passed through a sigmoid to obtain the video quality score $Q_v$ in $(0,1)$. Both scores are stored for each view and used jointly in the subsequent selection step.

Furthermore, selection operates on complete multi-view groups. Score thresholds are calibrated with manual inspection of representative samples, including samples near candidate decision boundaries, to account for differences in visual style. For a group $g$ with view set $\mathcal{V}_g$, we require
\begin{equation}
    \operatorname{keep}(g)
    =
    \bigwedge_{v\in\mathcal{V}_g}
    \left[
        H_v
        \land A_v \geq \tau_A^{(s)}
        \land Q_v \geq \tau_Q^{(s)}
    \right],
\end{equation}
where $H_v$ indicates that view $v$ passes the integrity and motion checks, and $\tau_A^{(s)}$ and $\tau_Q^{(s)}$ are the calibrated thresholds for source $s$. Any failing view excludes the entire group, preserving complete view sets in the selected data. Before export, we additionally verify agreement across views in temporal boundaries, frame counts, frame rates, and the shared caption.

\begin{figure*}[t]
    \centering
    \includegraphics[width=0.9775\textwidth]{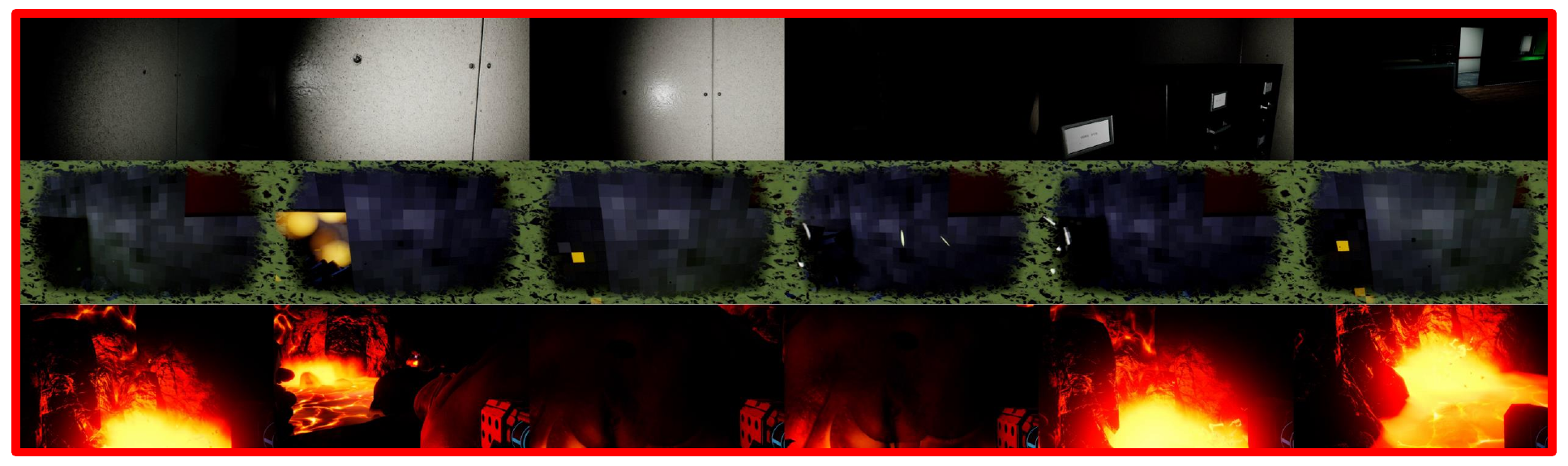}
    \caption{
    Examples of rejected video clips during quality filtering.
    Clips with severe underexposure or overexposure, abnormal in-game visual effects (e.g., poisoning), or substantial view occlusions are excluded due to degraded scene visibility and visual reliability.
    }
    \label{fig:data_false}
\end{figure*}

To qualitatively illustrate the filtering results, Figs.~\ref{fig:data_true} and~\ref{fig:data_false} present representative examples of retained and rejected video clips, respectively. The retained clips exhibit clear scene visibility, coherent temporal progression, and diverse visual content across different game environments. In contrast, rejected clips suffer from severe exposure abnormalities, visual distortions caused by abnormal in-game character states (e.g., poisoning), or substantial view occlusions that obscure scene content. These examples illustrate how the filtering pipeline removes visually degraded samples while preserving informative and temporally consistent observations for downstream spatial understanding and learning tasks.

\subsection{Data Annotation}
Unlike conventional video datasets that rely on vision-based algorithms to estimate geometric and motion annotations, GameRealm obtains ground-truth information directly from game engines through dedicated acquisition plugins. These plugins provide access to game-native information associated with game execution, enabling the collection of synchronized visual observations and corresponding geometric and control annotations.

This acquisition paradigm establishes direct correspondence between rendered observations and their underlying engine-provided information, including depth, camera parameters, player inputs, and 3D point trajectories. All annotations are associated through engine-provided timestamps, ensuring temporal synchronization across modalities. Furthermore, geometric annotations follow consistent metric units and coordinate conventions, facilitating unified spatial interpretation across different data modalities. Consequently, GameRealm provides engine-grounded supervision without relying on post-hoc visual estimation for these annotations. Textual descriptions are generated separately using vision-language models and are distinguished from game-native annotations.

\noindent\textbf{RGB observations.} RGB observations are organized by recording, segment, and camera, with multi-view videos sharing a common local frame index and retaining source-frame boundaries. Representative sequences have a resolution of \(1280\times720\), with corresponding depth maps, camera parameters, and available control inputs aligned to the exported timeline.

\noindent\textbf{Depth maps.}Depth annotations provide numerical depth information in multiple formats, accompanied by corresponding decoding rules and invalid-value conventions. Metric depth, coordinate alignment, pixel registration, and annotation validity are documented and aligned according to the specifications of each data release.

\noindent\textbf{Camera Parameter.}Camera annotations provide \(3\times3\) intrinsic matrices and \(4\times4\) camera-to-world transformations for supported sequences. Coordinate conventions, rotation composition, unit conversion, and synchronization information are documented for each data release, including explicit transformation

\noindent\textbf{Audio.}Audio annotations include 48 kHz, two-channel WAV recordings, together with associated temporal and segment-level metadata. Audio availability, completeness, and temporal synchronization are documented individually for each recording.

\noindent\textbf{Keypress.}Input annotations include keyboard and mouse states, such as directional keys, Space, and mouse buttons, with binary down and changed flags representing held states and state transitions. Where available, relative mouse displacements and frame-aligned timestamps are also retained to establish explicit temporal correspondence between user inputs and visual observations.

\noindent\textbf{Point tracking.} Point annotations represent persistent correspondences through triangle addresses and barycentric coordinates or vertex identities, providing world-space 3D positions and per-view 2D projections where available. Visibility, synchronization, point identity, and geometric validity are documented according to annotation readiness, with additional skeletal joints and skinned vertices available for selected hand sequences.

\begin{figure*}[b!]
\centering
  \includegraphics[width=0.95\textwidth]{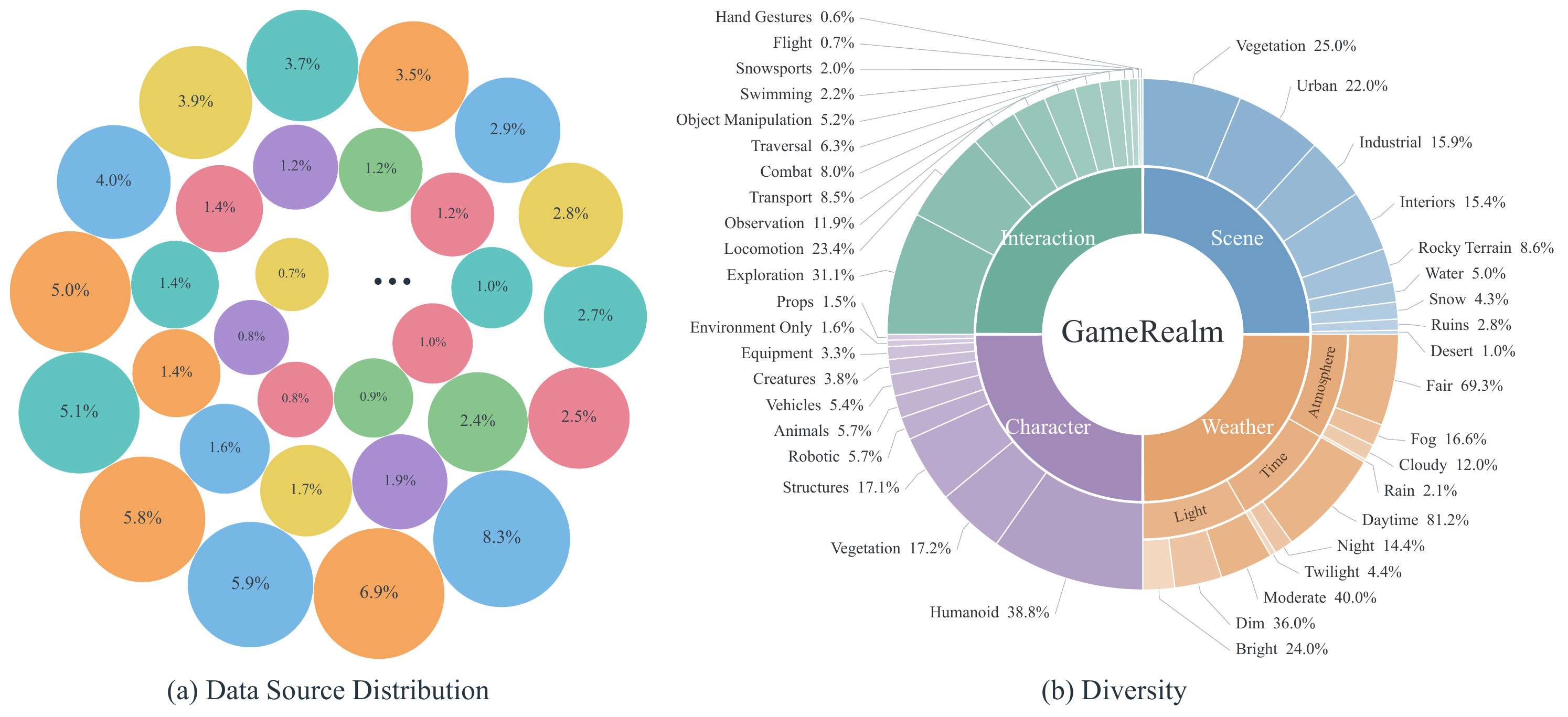}
  \caption{Data distribution}
  \label{fig:distribution}
\end{figure*}

\noindent\textbf{Video Caption.} We construct hierarchical textual annotations for synchronized multi-view gameplay clips using Qwen3.5-Flash~\citep{team2026qwen3}, as illustrated in Fig.~\ref{fig:Data_Acquisition}~(c). For each view, up to 200 frames are sampled after cropping and resizing, with original frame indices preserved. The model independently generates temporally ordered action descriptions covering entities, actions, spatial relationships, and state changes. These descriptions are concatenated into local captions and subsequently fused through a text-only call into a shared global caption, prioritizing the primary view while incorporating compatible information from other views. Both caption types are validated and retained separately for downstream textual conditioning.

\subsection{Data Statistics}

\noindent\textbf{Data scale and modality coverage.} GameRealm contains over 120 million RGB frames. Following Syn4D~\citep{jiang2026syn4d} and SynthVerse~\citep{Zhao2026SynthVerse}, we count synchronized frames from different cameras separately. Based on typical recording rates of 16--24~fps, we estimate over 1{,}000 hours of video summed across all views. Each RGB frame is paired with a metric depth map and camera parameters. Multi-view and panoramic sequences also include synchronized audio and action annotations, providing multimodal supervision for interactive world modeling. Fig.~\ref{fig:distribution} shows the frame distribution across more than 50 games. The number of frames collected per game depends on the size of its playable maps, with larger game worlds generally contributing more frames.

\noindent\textbf{Data composition.}  The dataset is predominantly composed of multi-view data (77.6\%), followed by stereo (9.2\%), panoramic (7.5\%), and single-view data (5.7\%). The dominance of multi-view data reflects our acquisition design, which prioritizes synchronized multi-camera capture to provide broader spatial coverage and richer geometric supervision across diverse game environments. Stereo, panoramic, and single-view data are collected as complementary configurations for specific visual learning tasks. Within the tracking subset, multi-view and single-view observations account for 63.9\% and 36.1\%, respectively. Among samples with explicit perspective annotations, first-person and third-person views constitute 45.6\% and 54.4\%, respectively, indicating a relatively balanced representation of the two perspectives.
\begin{figure*}[t]
\centering
  \includegraphics[width=0.99\textwidth]{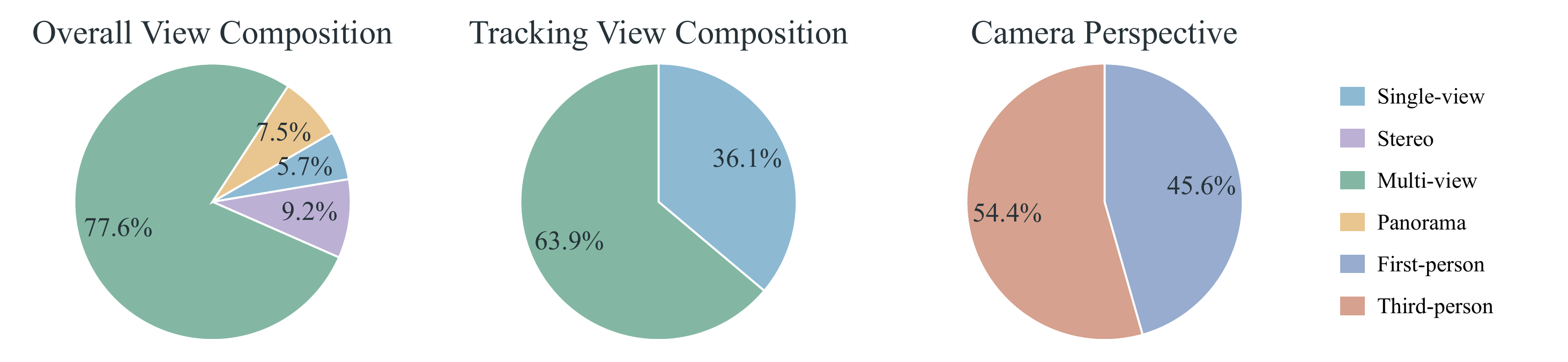}
  \caption{Data composition}
  \label{fig:data_composition}
\end{figure*}

\begin{figure*}[h]
\centering
  \includegraphics[width=0.9775\textwidth]{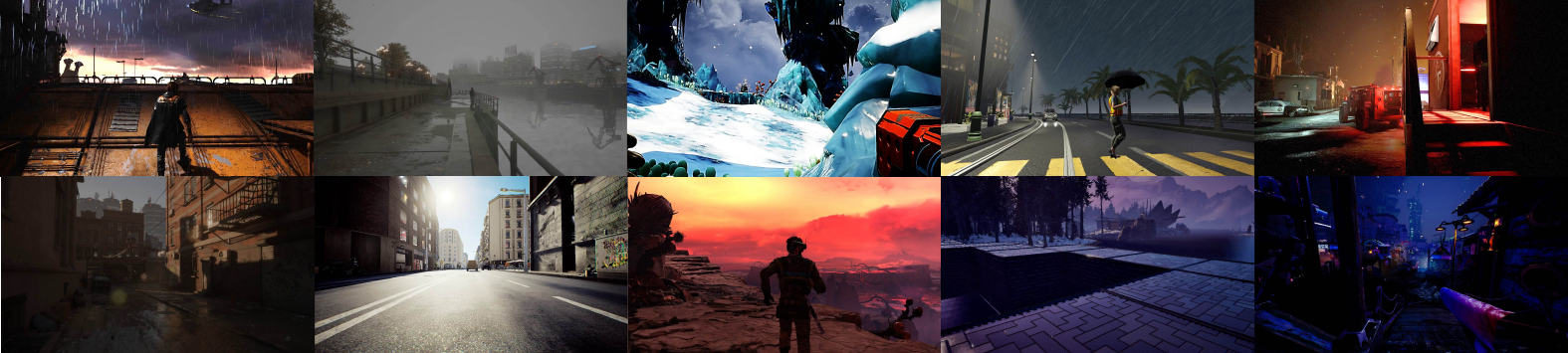}
  \caption{Weather. The first row illustrates weather conditions, including rain, snow, and fog, while the second row shows the progression of time from pre-dawn and morning through noon, dusk, and late night. }
  \label{fig:Weather}
\end{figure*}

\begin{figure*}[h]
  \includegraphics[width=0.9775\textwidth]{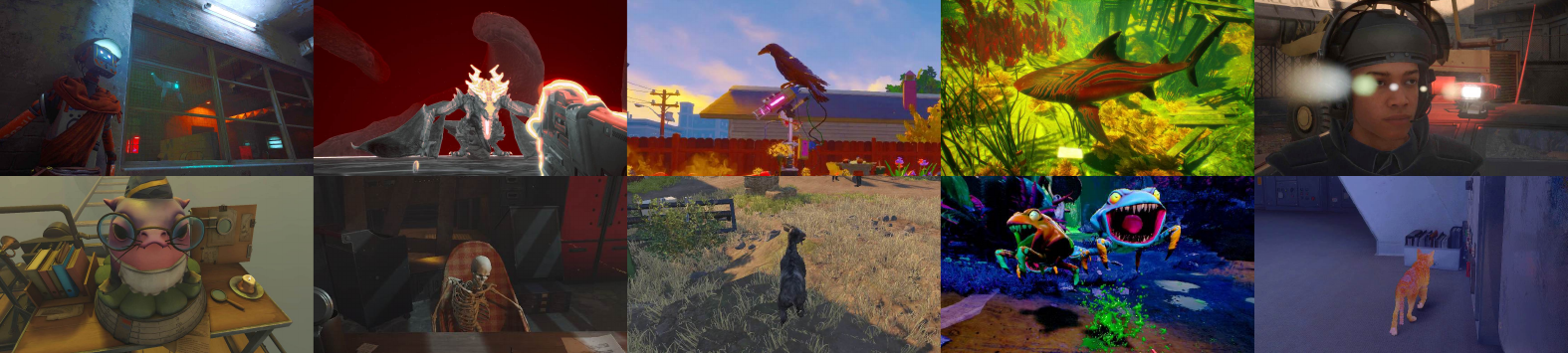}
  \caption{Character. Representative examples of character diversity in GameRealm, including robots, fish, birds, dragons, humans, and many others.}
  \label{fig:Character}
\end{figure*}

GameRealm encompasses diverse game content across four semantic dimensions: scene, weather, character, and interaction, as summarized in Fig.~\ref{fig:distribution} . These dimensions capture complementary aspects of gameplay, including environmental conditions, spatial settings, agent behaviors, and entity types. Fig. ~\ref{fig:Weather}--\ref{fig:Character} present representative examples of these categories, illustrating the diversity of visual content covered by our data acquisition pipeline.

\begin{figure*}[t]
\centering
  \includegraphics[width=0.9775\textwidth]{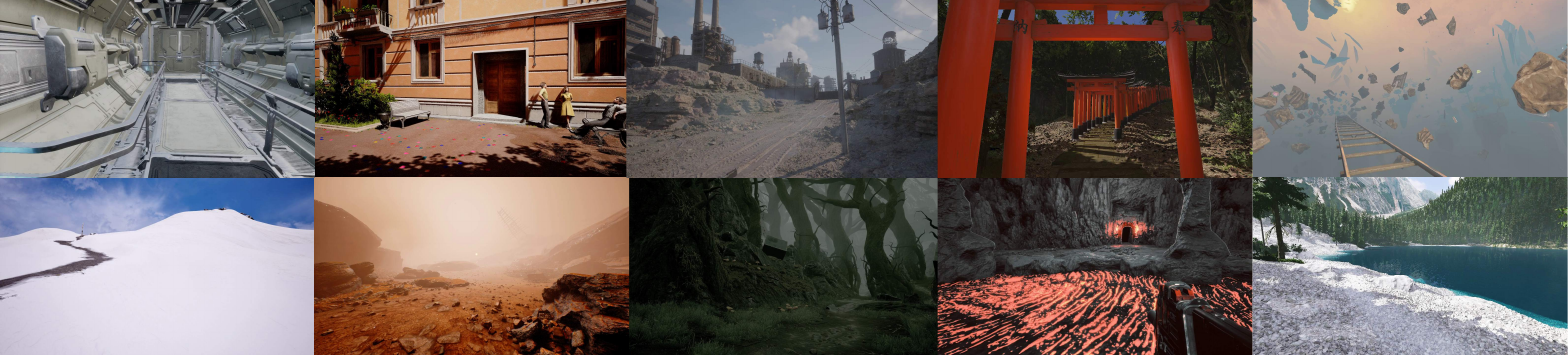}
  \caption{Scene. The first row shows human-oriented environments, including indoor spaces, urban streets, and outer space, while the second row presents natural environments such as snowfields, forests, lava fields, and lakes.}
  \label{fig:Scene}
\end{figure*}

\noindent\textbf{Scene diversity.} Figure~\ref{fig:Scene} illustrates the broad range of environments represented in GameRealm, spanning natural scenes such as forests, grasslands, mountains, deserts, rivers, lakes, beaches, underwater environments, and caves, as well as human-made settings such as urban streets, villages, and laboratories. This diversity encompasses different spatial layouts, geometric structures, and visual characteristics, enabling the study of scene understanding and geometric learning across heterogeneous environments.

\noindent\textbf{Weather diversity.} As shown in Fig.~\ref{fig:Weather}, GameRealm covers a variety of weather conditions and times of day, including sunny, cloudy, rainy, snowy, foggy, and stormy environments, as well as variations in daylight and illumination. These conditions introduce visual changes in scene appearance, lighting, visibility, and atmospheric effects, providing diverse environmental contexts for learning robust visual representations and modeling dynamic worlds.

\noindent\textbf{Character diversity.} Figure~\ref{fig:Character} demonstrates the diversity of entities appearing in GameRealm, including humans, robots, cats, canines, birds, sharks, ants, beetles, spiders, and monsters. These entities exhibit different appearances, shapes, scales, and motion characteristics, broadening the range of visual and dynamic patterns represented in the dataset. Combined with the diversity of scenes and interactions, this variety supports the evaluation of world models and visual understanding methods across different characters and gameplay contexts.

\begin{figure*}[h]
\centering
  \includegraphics[width=0.9775\textwidth]{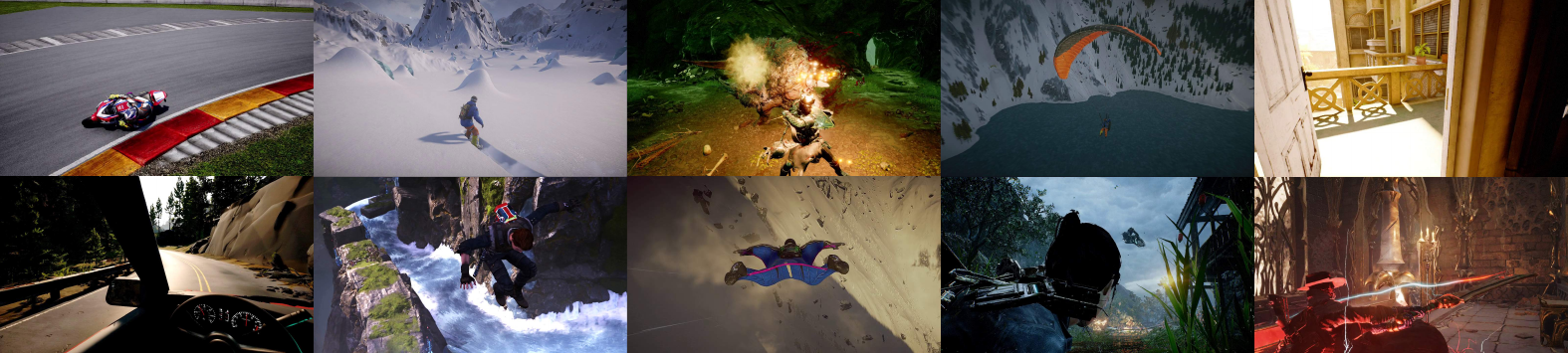}
  \caption{Interaction. Representative interaction activities in GameRealm, including motorcycle riding, driving, skiing, opening doors, jumping, and other diverse gameplay interactions.}
  \label{fig:Interaction}
\end{figure*}

\noindent\textbf{Interaction diversity.} As illustrated in Fig.~\ref{fig:Interaction}, GameRealm captures a wide range of gameplay activities, including roaming, walking, running, jumping, climbing, swimming, aiming, shooting, blocking, dodging, driving, riding, boating, skiing, and object-related operations such as picking, placing, opening, closing, folding, and removing. These activities introduce varied motion patterns and interaction dynamics, supporting the investigation of action-conditioned world modeling and the representation of dynamic scenes under different control inputs.

\begin{figure*}[b!]
    \centering
    \includegraphics[width=0.9775\textwidth]{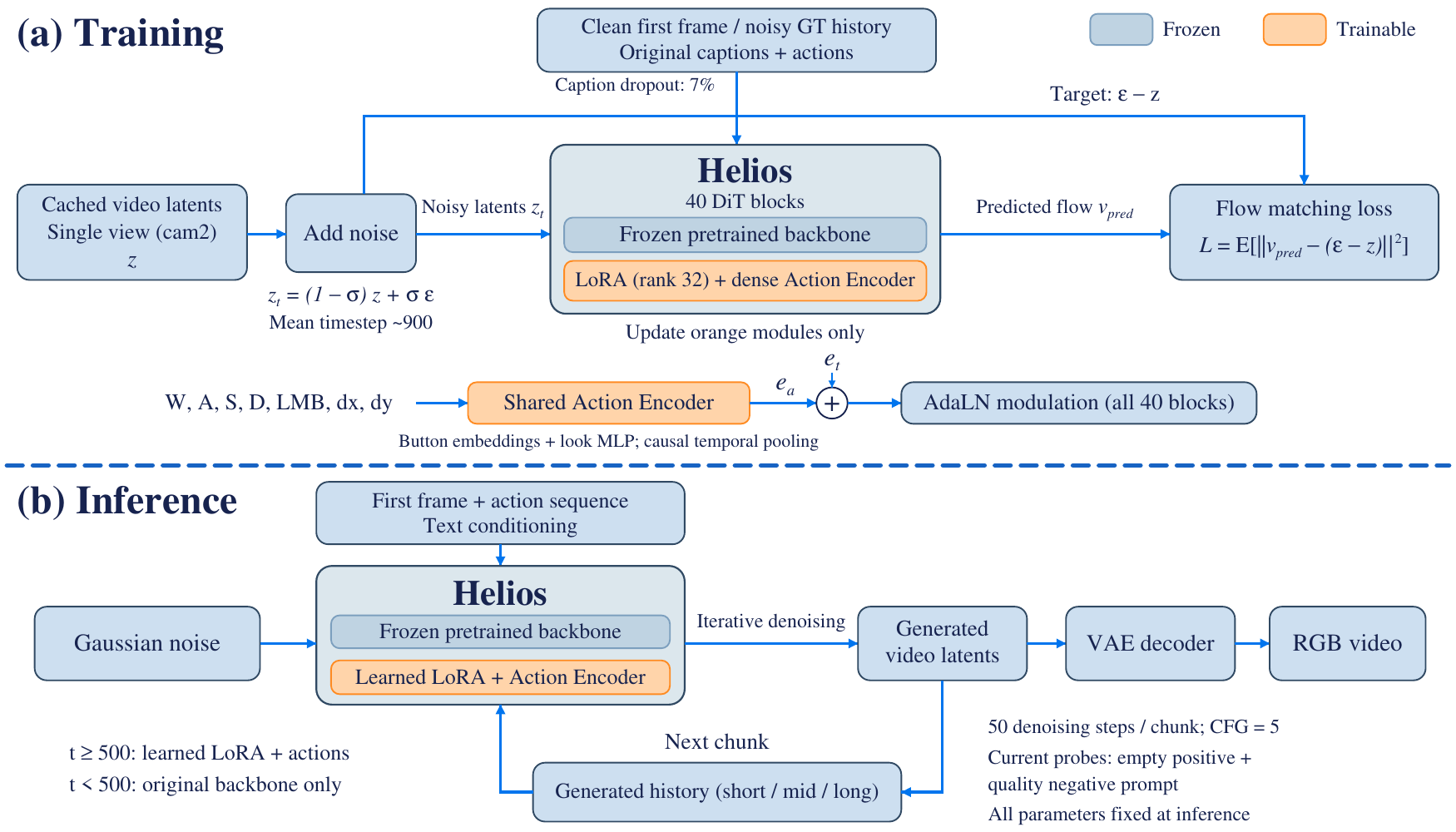}
    \caption{Training and inference of action-conditioned Helios~\citep{yuan2026helios}. (a) Parameter-efficient adaptation with LoRA and action conditioning. (b) Autoregressive video generation with hybrid denoising. Orange denotes trainable components.}
    \label{fig:helios-training-inference}
\end{figure*}

\section{Interactive World Model}
\label{sec:IWM}

We evaluate GameRealm for interactive world modeling by adapting pretrained generative models with LoRA~\citep{hu2021lora} and lightweight conditioning modules. We investigate three settings: single-view action-conditioned generation, synchronized multi-view generation, and panoramic audio-visual generation.

\subsection{Single-View World Model}
\label{sec:SWM}

\noindent\textbf{Network architecture.}
As illustrated in Fig.~\ref{fig:helios-training-inference}, we adapt the pretrained 40-block Helios backbone. The backbone and video VAE remain frozen, while rank-32 LoRA adapters~\citep{hu2021lora} and an action encoder are optimized. Each frame is associated with a seven-dimensional action vector
\begin{equation}
    \mathbf{u}_i =
    (w_i,a_i,s_i,d_i,b_i,\Delta x_i,\Delta y_i),
\end{equation}
where $w_i,a_i,s_i,d_i$ represent directional key states, $b_i$ denotes the left mouse button state, and $\Delta x_i,\Delta y_i$ encode mouse displacements. The action encoder combines learned discrete embeddings with an MLP for continuous controls. For each 33-frame training window, action features are temporally aggregated into nine embeddings aligned with the video latent frame rate. These features are injected into the diffusion-timestep embeddings and used to modulate all transformer blocks through adaptive layer normalization.

\noindent\textbf{Flow-matching training.}
Let $\mathbf{z}_0$ denote the clean target video latents and $\mathcal{C}$ the conditioning information, including the first-frame latent, historical latents, and text embedding. We use teacher-forced ground-truth history, perturbing historical latents while keeping the first frame clean. Given Gaussian noise $\boldsymbol{\epsilon}\sim\mathcal{N}(\mathbf{0},\mathbf{I})$, the noisy target and target flow are
\begin{equation}
    \begin{aligned}
        \mathbf{z}_t
        &= (1-\sigma_t)\mathbf{z}_0
        +\sigma_t\boldsymbol{\epsilon},\\
        \mathbf{v}^{\star}
        &=\boldsymbol{\epsilon}-\mathbf{z}_0,
    \end{aligned}
    \label{eq:helios-flow-target}
\end{equation}
where $\sigma_t$ denotes the noise level at timestep $t$. The model predicts
\begin{equation}
    \widehat{\mathbf{v}}_t =
    f_{\boldsymbol{\theta}_0,\boldsymbol{\phi},\boldsymbol{\psi}}
    (\mathbf{z}_t,t,\mathcal{C},\mathbf{u}),
\end{equation}
where $\boldsymbol{\theta}_0$, $\boldsymbol{\phi}$, and $\boldsymbol{\psi}$ denote the frozen backbone, LoRA, and action-encoder parameters, respectively. We optimize the flow-matching objective
\begin{equation}
    \mathcal{L}_{\mathrm{FM}}
    =
    \mathbb{E}\left[
    \frac{1}{N}
    \left\|
    \widehat{\mathbf{v}}_t-\mathbf{v}^{\star}
    \right\|_2^2
    \right],
    \label{eq:helios-flow-loss}
\end{equation}
where $N$ is the number of target latent elements. Only target latents contribute to the loss, and only $\boldsymbol{\phi}$ and $\boldsymbol{\psi}$ are updated. We use an encoded empty text prompt throughout training.

Timesteps are sampled from a logit-normal distribution with underlying Gaussian parameters $\mu=-1.651877$ and standard deviation $1$, mapped to the descending Helios noise schedule.

\noindent\textbf{Autoregressive inference.}
Given an initial frame and an action sequence, the model generates video chunks from Gaussian noise using 50 denoising steps. We adopt a hybrid prediction strategy:
\begin{equation}
    \widehat{\mathbf{v}}_t =
    \begin{cases}
    f_{\boldsymbol{\theta}_0,\boldsymbol{\phi},\boldsymbol{\psi}}
    (\mathbf{z}_t,t,\mathcal{C},\mathbf{u}),
    & t\geq 500,\\
    f_{\boldsymbol{\theta}_0}
    (\mathbf{z}_t,t,\mathcal{C}),
    & t<500.
    \end{cases}
    \label{eq:helios-hybrid-inference}
\end{equation}
Thus, the learned LoRA and action conditioning are active only at higher noise levels. After denoising, the frozen VAE decodes the generated latents into RGB frames, which are appended to the history for subsequent generation. All parameters remain fixed during inference.

\noindent\textbf{Implementation details.}
We train at $384\times640$ resolution with a global batch size of 128 across 64 GPUs. LoRA and the action encoder are optimized using AdamW with a learning rate of $10^{-4}$, 100 warmup steps, and a constant learning rate for 20,000 optimizer steps.

\begin{figure*}[p]
  \centering

  \includegraphics[width=0.9775\textwidth, height=0.28\textheight,
                   keepaspectratio]{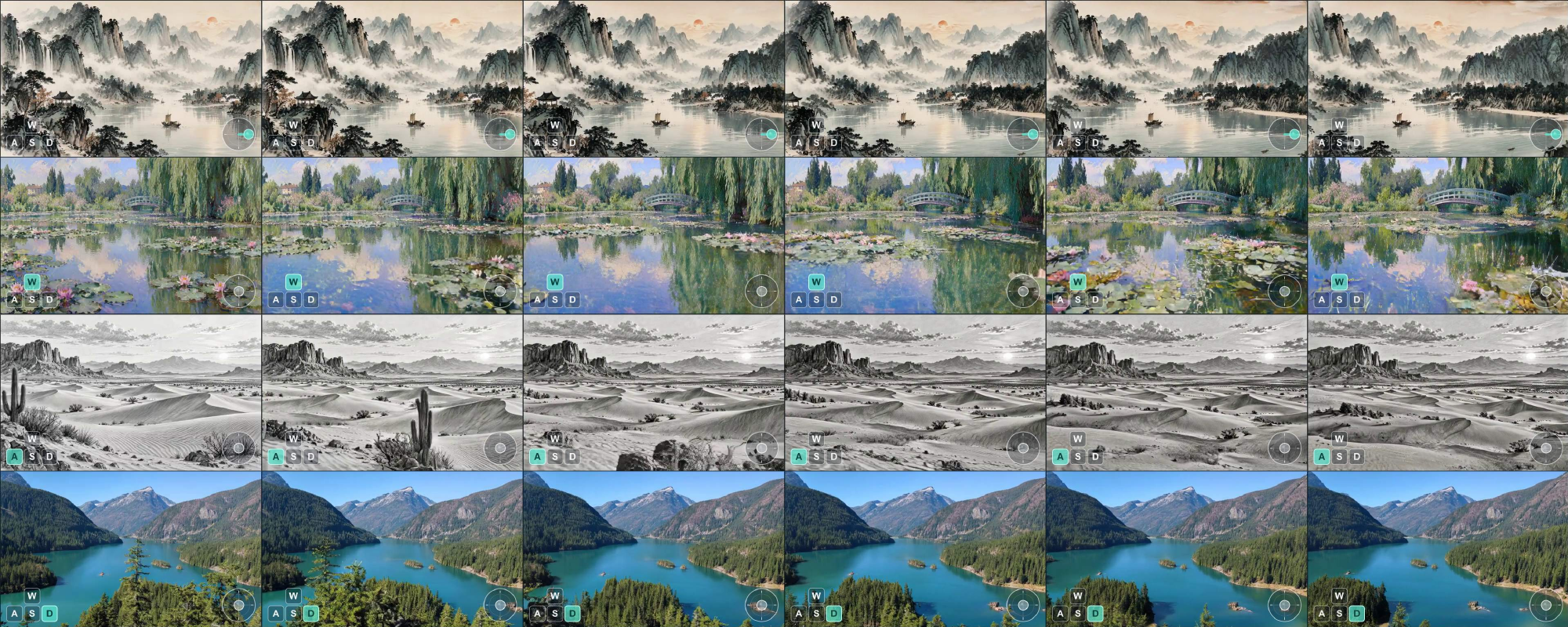}
  \caption{Interactive world model performance with different image styles}
  \label{fig:Style}

  \medskip

  \includegraphics[width=0.9775\textwidth, height=0.38\textheight,
                   keepaspectratio]{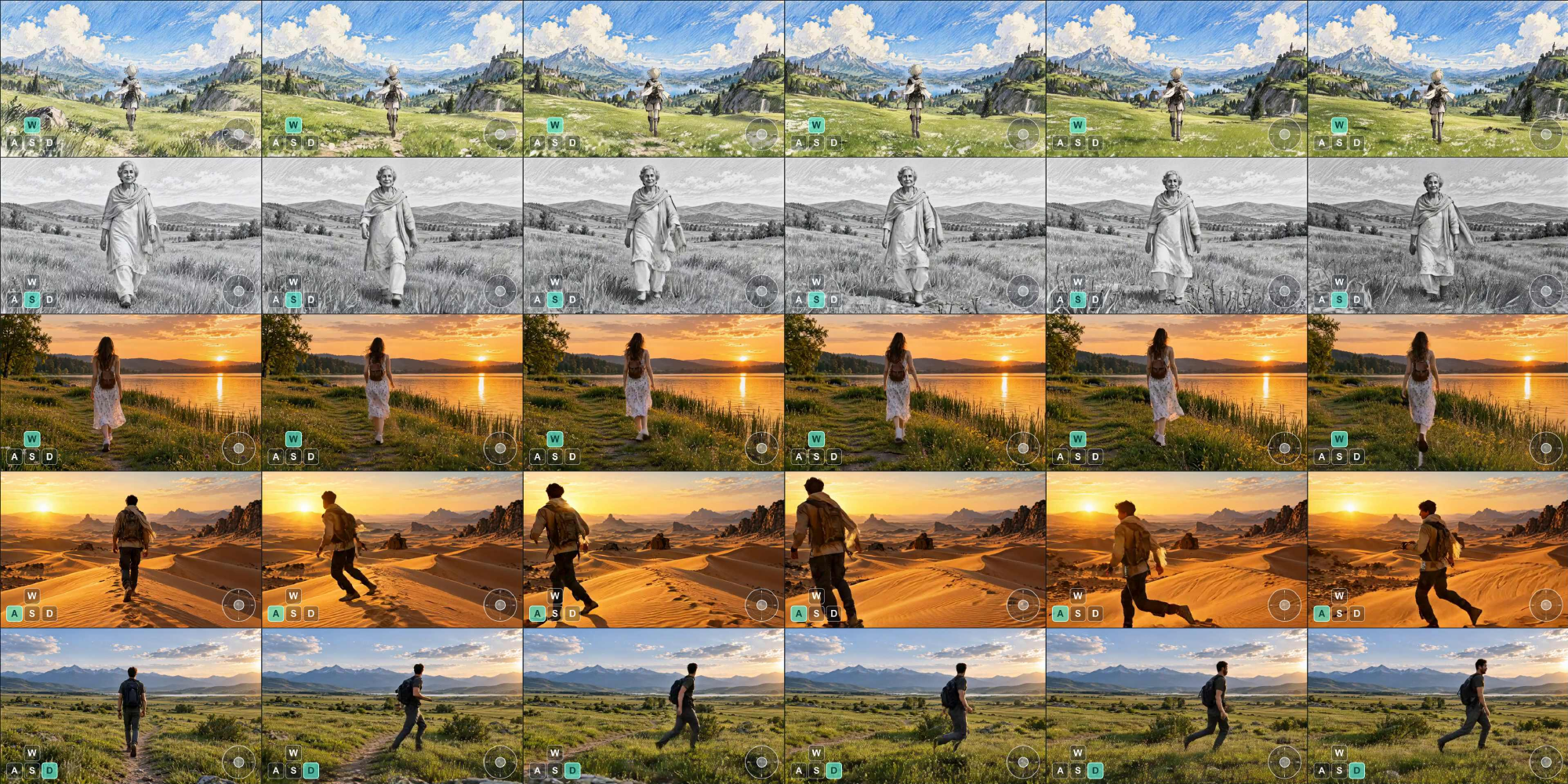}
  \caption{Interactive world model performance in third-person scenarios}
  \label{fig:third}

  \medskip

  \includegraphics[width=0.9775\textwidth, height=0.33\textheight,
                   keepaspectratio]{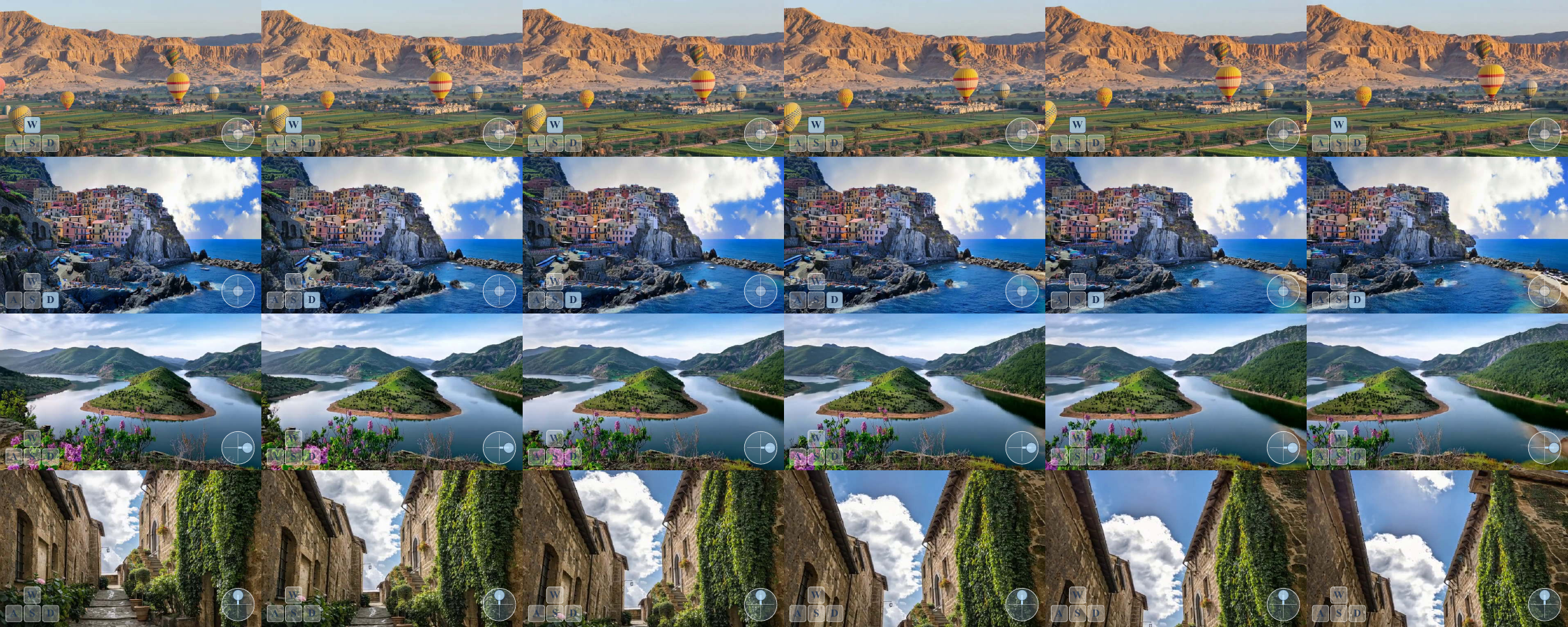}
  \caption{Interactive Generation on VBench++. From top to bottom:
    forward translation, rightward translation, rightward rotation,
    and upward rotation. Each row shows the input image followed by
    five generated frames.}
  \label{fig:vbench}
\end{figure*}

\begin{figure*}[ht]
  \includegraphics[width=0.9775\textwidth]{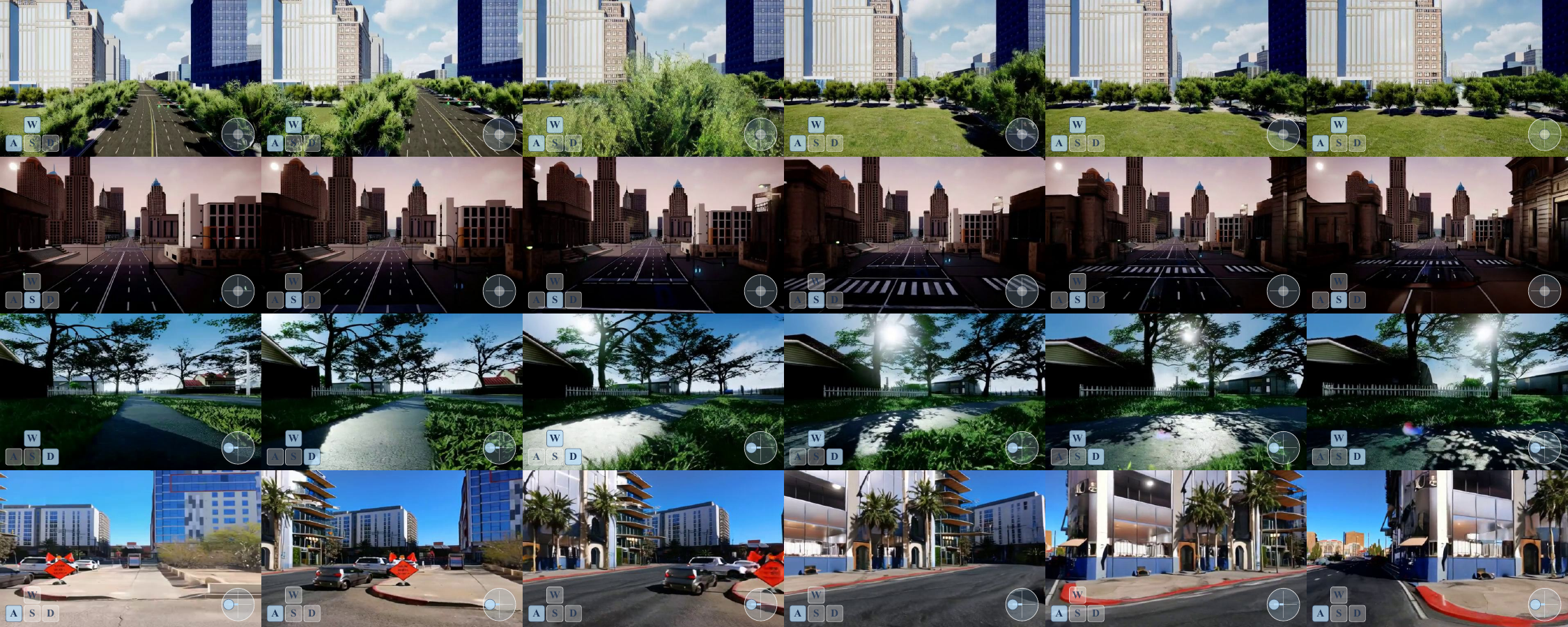}
  \caption{Interactive generation on IWorld-Bench, illustrating single-key movement, combined-key movement, and joint translation--rotation control.}
  \label{fig:iworld}
\end{figure*}

\noindent\textbf{Quantitative results}
We evaluate the effectiveness of GameRealm's action annotations for interactive world modeling using VBench++~\citep{huang2025vbench++} and iWorld-Bench~\citep{fang2026iworld}. As shown in Tab.~\ref{tab:stationary}, the fine-tuned Helios maintains visual quality and motion stability under zero-action conditions. We further assess directional controllability across eight inputs, including WASD movement and joystick-based viewpoint control (up, down, left, and right). As shown in Tab.~\ref{tab:motion_alignment}, Helios$^{*}$ achieves a direction success rate of 76.59\%, demonstrating that GameRealm's action annotations provide effective supervision for learning controllable visual dynamics. The original text-conditioned Helios serves as a reference rather than a matched action-conditioned baseline.  Figs.~\ref{fig:vbench} and~\ref{fig:iworld} present qualitative visualizations on VBench and iWorld-Bench.

\begin{table*}[ht]
    \centering

    \caption{Visual quality and motion stability under different conditioning settings.}

    \label{tab:stationary}
    \small
    \setlength{\tabcolsep}{7pt}
    \renewcommand{\arraystretch}{1.0}
    \resizebox{\textwidth}{!}{%
    \begin{tabular}{ll ccc ccc}
        \toprule
        & & \multicolumn{3}{c}{\textbf{VBench-I2V~\citep{huang2025vbench++}}}
          & \multicolumn{3}{c}{\textbf{iWorld-Bench~\citep{fang2026iworld}}} \\
        \cmidrule(lr){3-5}\cmidrule(lr){6-8}
        Method & Condition
        & MUSIQ $\uparrow$
        & Rotation path ($^\circ$)$\downarrow$
        & Endpoint rotation ($^\circ$)$\downarrow$
        & MUSIQ $\uparrow$
        & Rotation path ($^\circ$)$\downarrow$
        & Endpoint rotation ($^\circ$)$\downarrow$ \\
        \midrule
        Helios & Empty text
        & 0.6732 & 5.1107 & 1.4688
        & 0.5504 & 4.1446 & 1.7275 \\

        Helios & Stationary text
        & 0.6879 & 3.4294 & 0.6952
        & 0.5701 & 0.3140 & 0.0735 \\

        Helios$^{*}$ & Zero action
        & \textbf{0.6949} & \textbf{3.3018} & \textbf{0.6421}
        & \textbf{0.5846} & \textbf{0.2835} & \textbf{0.0461} \\
        \bottomrule
    \end{tabular}%
    }
\end{table*}

\begin{table}[ht]
    \centering
    \caption{Directional controllability enabled by GameRealm action supervision. Helios$^{*}$ is the model fine-tuned on GameRealm, while the original text-conditioned Helios is included for reference.}
    \label{tab:motion_alignment}
    \scriptsize
    \setlength{\tabcolsep}{2pt}
    \renewcommand{\arraystretch}{1.0}
    \resizebox{0.55\columnwidth}{!}{%
    \begin{tabular}{@{}llccc@{}}
        \toprule
        Method & Condition
        & Trans. cos. $\uparrow$
        & Rot. cos. $\uparrow$
        & Dir. succ. (\%) $\uparrow$ \\
        \midrule
        Helios & Empty text
        & 0.0020 & 0.0005 & 1.52 \\
        Helios& Directional text
        & 0.0015 & 0.0047 & 2.40 \\
        Helios$^{*}$ & Action
        & \textbf{0.5490} & \textbf{0.8685}
        & \textbf{76.59} \\
        \bottomrule
    \end{tabular}%
    }
\end{table}

\begin{figure*}[b!]
  \includegraphics[width=0.9775\textwidth]{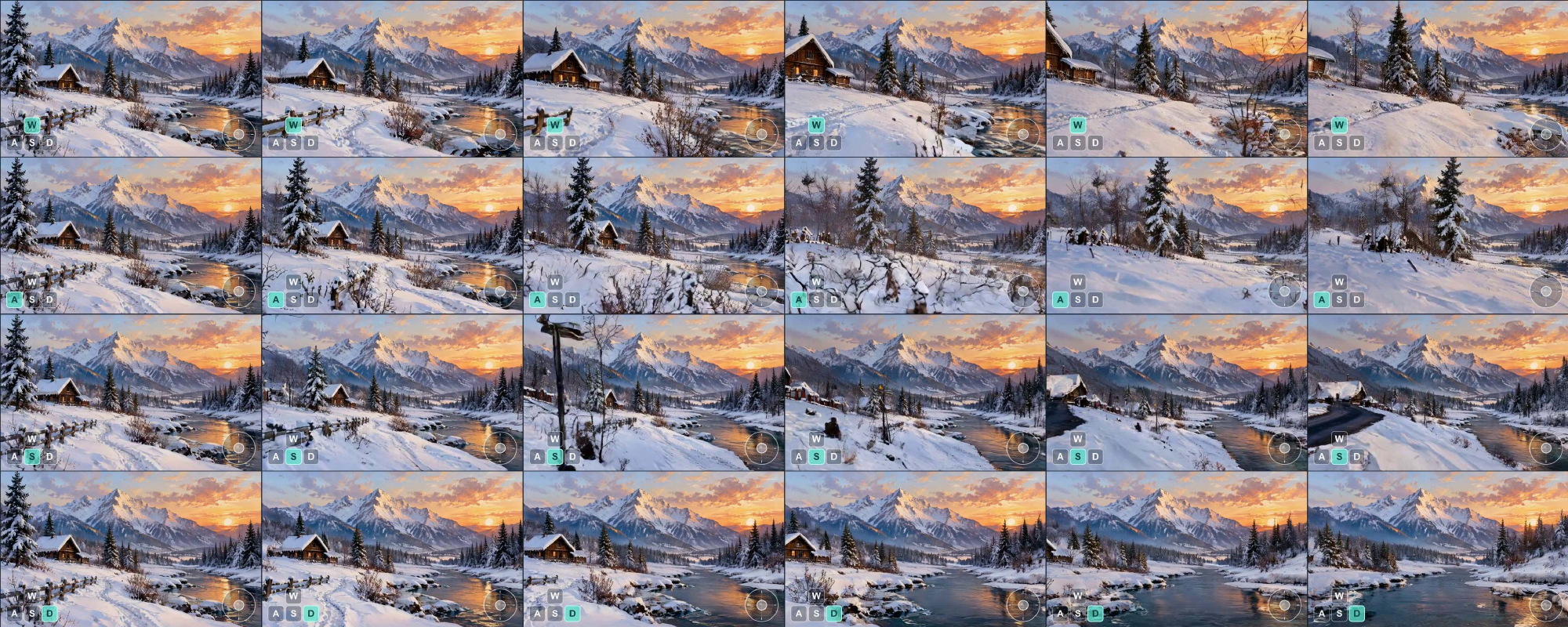}
  \caption{Interactive World Model Performance with WASD Control}
  \label{fig:Snow_wasd}
\end{figure*}

To further validate the effectiveness of GameRealm's action supervision, we vary the input actions of Helios$^{*}$ while keeping the model and initial observations fixed on iWorld-Bench. As shown in Tab.~\ref{tab:action_ablation}, correct actions achieve a translation cosine similarity of 0.7783, compared with -0.0015 for zero actions and -0.7997 for reversed actions. MUSIQ scores remain comparable across conditions. These results demonstrate that generated motion responds consistently to the supplied actions, supporting the effectiveness of GameRealm's action annotations for interactive world models.

\begin{table}[ht]
\centering
    \caption{Effect of action perturbations on generated motion and visual quality.}
    \label{tab:action_ablation}
    \small
    \setlength{\tabcolsep}{6pt}
    \renewcommand{\arraystretch}{1.0}
    \begin{tabular}{lcc}
        \toprule
        Condition
        & Translation Cosine $\uparrow$
        & MUSIQ $\uparrow$ \\
        \midrule
        Correct action
        & \textbf{0.7783} & 0.5706 \\
        Zero action
        & -0.0015 & \textbf{0.5867} \\
        Reversed action
        & -0.7997 & 0.5674 \\
        \bottomrule
    \end{tabular}
\end{table}

\begin{figure*}[ht]
  \includegraphics[width=0.9775\textwidth]{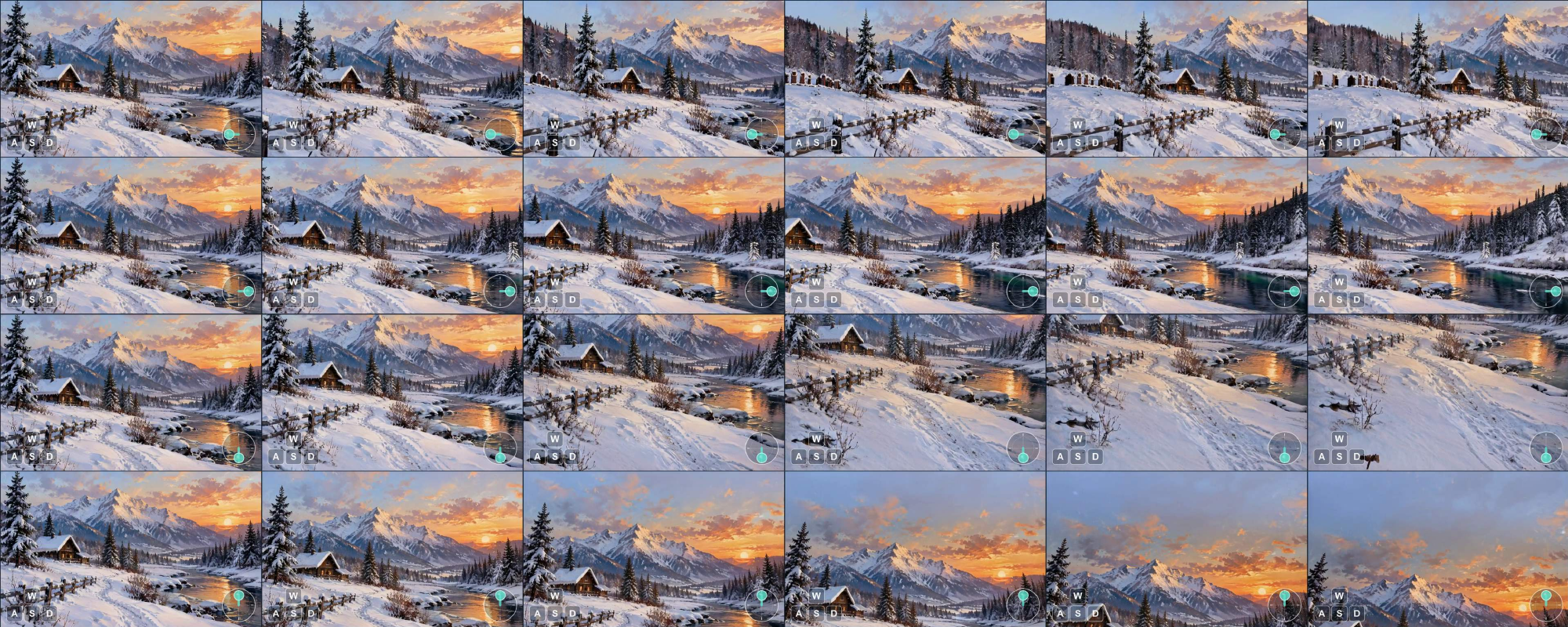}
  \caption{Interactive World Model Performance with Joystick View Control}
  \label{fig:Snow_dxdy}
\end{figure*}

\begin{figure*}[ht]
  \includegraphics[width=0.9775\textwidth]{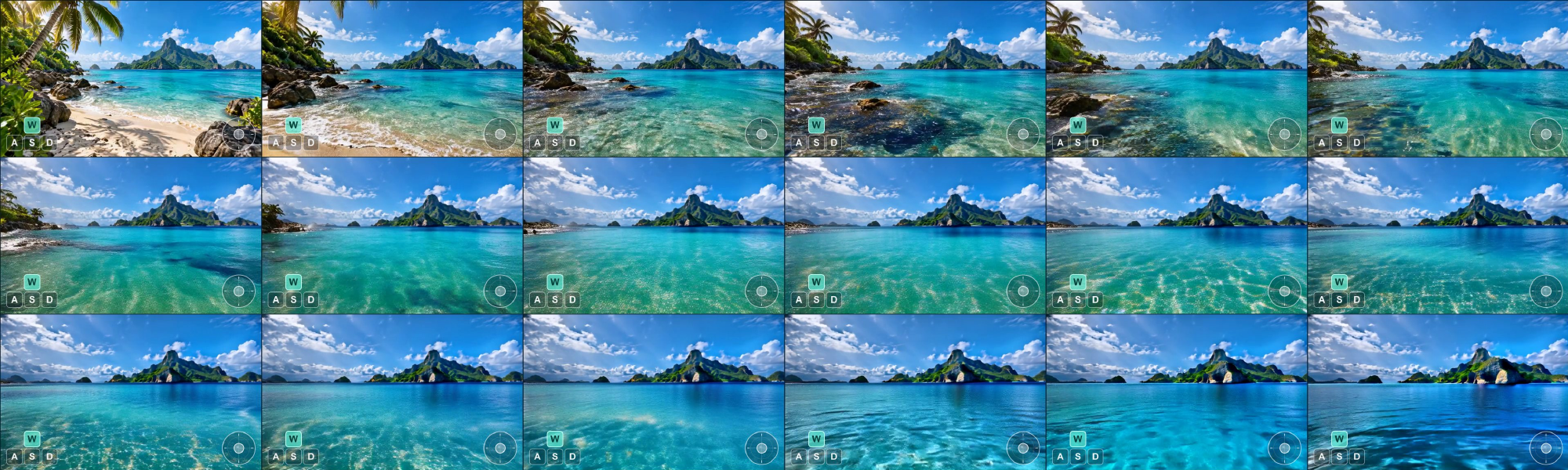}
  \caption{Interactive World Model Performance with Long-horizon Generation}
  \label{fig:IWM_long}
\end{figure*}

\noindent\textbf{Qualitative results.}
Figs.~\ref{fig:Snow_wasd} and~\ref{fig:Snow_dxdy} illustrate how the model responds to keyboard and mouse inputs, respectively. Keyboard commands steer the character through the scene, while mouse movements adjust the camera orientation and bring different parts of the environment into view. Over the longer rollout shown in Fig.~\ref{fig:IWM_long}, scene elements remain recognizable as the viewpoint changes, and their overall spatial arrangement remains visually consistent.

Fig.~\ref{fig:Style} further shows that, although our training data are derived entirely from interactive gameplay, the resulting model generalizes effectively to videos with diverse visual styles, including ink painting, oil painting, pencil sketch, and realistic photography. Fig.~\ref{fig:third} further demonstrates that GameRealm supports interactive generation in third-person scenarios across diverse character appearances and visual styles. By keyboard inputs, users can control character movements, such as walking, demonstrating the applicability of our action-conditioned model beyond first-person gameplay.  Additional evaluations on VBench++~\ref{fig:vbench} and IWorld-Bench~\ref{fig:iworld}  demonstrate its ability to follow translation and rotation commands, including combined control inputs.

\subsection{Multi-view World Model}
\label{sec:multiview-world-model}

\begin{figure*}[b!]
  \includegraphics[width=0.9775\textwidth]{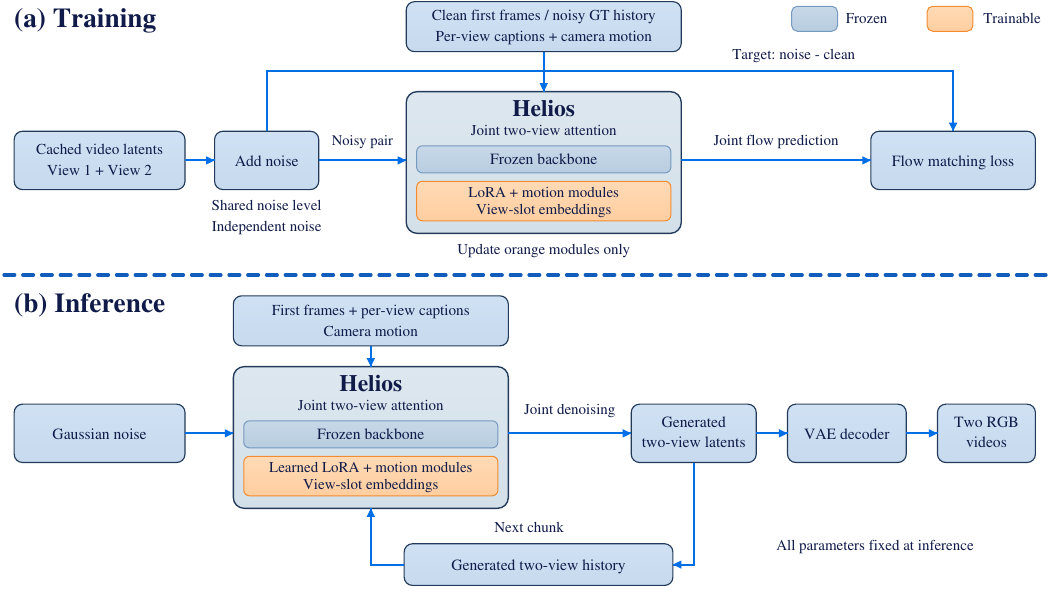}
  \caption{Training and inference of the multi-view world model. During training, Helios jointly predicts flow velocities for synchronized views conditioned on first frames, ground-truth history, captions, and camera motion. During inference, joint denoising and autoregressive history updates generate synchronized video streams.}
  \label{fig:multiview_pipeline}
\end{figure*}

\noindent\textbf{Network architecture.}
As shown in Fig.~\ref{fig:multiview_pipeline}, we extend the adapted Helios model to jointly generate synchronized views. Each training sample contains two views from the same recording group and temporal window. Their visual tokens are concatenated into a joint sequence, enabling cross-view information exchange through self-attention. View-slot embeddings distinguish the two streams, while view-specific captions provide textual conditioning.

Camera motion is encoded using per-view trajectories and camera intrinsics. We first calibrate a velocity threshold for each game by aligning camera-derived lateral and forward motion with recorded W/A/S/D inputs from the primary view. The calibrated threshold is then applied to each view to obtain directional motion states, while relative rotations remain continuous. These motion conditions are converted into per-pixel origin--direction features and injected into the transformer through adaptive scale and shift modulation. The pretrained backbone and VAE remain frozen; only rank-32 LoRA adapters, motion-conditioning modules, and view-slot embeddings are optimized.

\noindent\textbf{Flow-matching training.}
We follow the single-view formulation in Eq.~\eqref{eq:helios-flow-target}, applying the same noise level to both views while independently sampling their Gaussian noise. Let $\mathbf{z}_0^{(k)}$ and $\boldsymbol{\epsilon}^{(k)}$ denote the clean latents and noise for view $k\in\{1,2\}$. The noisy latents are given by
\begin{equation}
    \mathbf{z}_t^{(k)}
    =
    (1-\sigma_t)\mathbf{z}_0^{(k)}
    +\sigma_t\boldsymbol{\epsilon}^{(k)}.
\end{equation}
The conditioning information $\mathcal{C}$ includes the first-frame latents, historical latents, view-specific text embeddings, and camera-motion features. The model jointly predicts both flow fields and minimizes
\begin{equation}
    \mathcal{L}_{\mathrm{MV}}
    =
    \mathbb{E}\left[
    \frac{1}{2}
    \sum_{k=1}^{2}
    \frac{1}{N_k}
    \left\|
    f_{\Theta}^{(k)}
    (\mathbf{z}_t^{(1)},\mathbf{z}_t^{(2)},t,\mathcal{C})
    -
    (\boldsymbol{\epsilon}^{(k)}-\mathbf{z}_0^{(k)})
    \right\|_2^2
    \right],
    \label{eq:mv-flow-matching}
\end{equation}
where $f_{\Theta}^{(k)}$ denotes the predicted flow for view $k$, and $N_k$ is the number of target latent elements in that view. We use teacher-forced ground-truth history, perturbing historical latents while keeping the first frames clean. Cross-view dependencies are learned through joint attention without an additional explicit geometric consistency loss. The timestep distribution follows the single-view sampling procedure, with the underlying Gaussian mean set to $-2$ and standard deviation set to $1$.

\noindent\textbf{Autoregressive inference.}
Given two initial frames, their captions, and camera-motion sequences, both views are jointly generated from Gaussian noise through iterative denoising. Generated chunks are appended to their respective histories and jointly condition subsequent predictions. This preserves cross-view information exchange throughout the rollout. The frozen VAE decodes the resulting latents into synchronized RGB videos, with all model parameters fixed during inference.

We evaluate cross-view geometric consistency by comparing the relative camera rotation estimated from the two generated views at each timestamp with the corresponding ground truth. We compare two models trained on the same synchronized multi-view data from the same initialization, with cross-view communication disabled (B1) or enabled (B2). As shown in Tab.~\ref{tab:cross_view_communication}, enabling cross-view communication reduces the inter-view rotation error from $39.93^\circ$ to $19.80^\circ$ across 88 evaluation cases, a relative reduction of 50.4\%. Rotation and translation errors also decrease when Pi3X~\citep{wang2026pi} jointly processes both views and each trajectory is evaluated after independent alignment. Results indicate that cross-view communication helps the model preserve the relative camera orientation across generated views.

\begin{table}[ht]
    \centering
    \caption{Cross-view communication ablation on test cases.}
    \label{tab:cross_view_communication}
    \small
    \setlength{\tabcolsep}{3pt}
    \begin{tabular}{@{}lccc@{}}
        \toprule
        Cross-view comm. & Inter-view $R$ ($^\circ$) $\downarrow$ & Joint $R$ ($^\circ$) $\downarrow$ & Joint $T$ $\downarrow$ \\
        \midrule
        Disabled (B1) & 39.93 & 13.94 & 2.88 \\
        Enabled (B2)  & \textbf{19.80} & \textbf{11.43} & \textbf{2.22} \\
        \bottomrule
    \end{tabular}
\end{table}

\begin{figure*}[ht]
  \includegraphics[width=0.9775\textwidth]{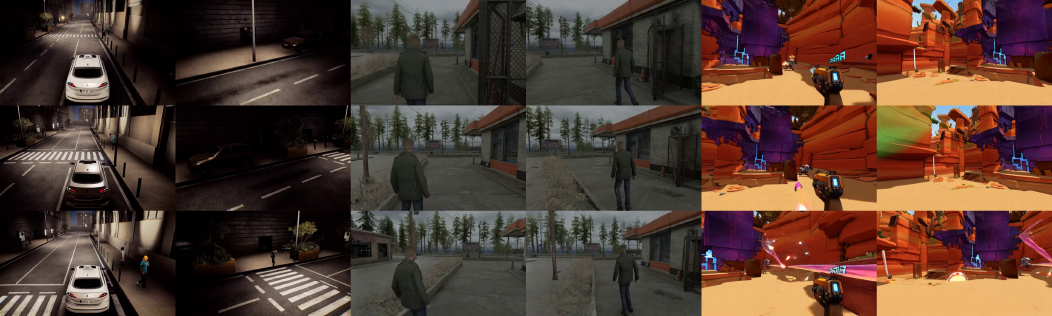}
  \caption{Multi-View world model generation. Each column represents a different viewpoint.}
  \label{fig:multiview_demo}
\end{figure*}

\subsection{Panoramic Audio-Visual World Model}
\label{sec:panoramic-world-model}

\begin{figure*}[b!]
    \centering
    \includegraphics[width=0.9775\textwidth]{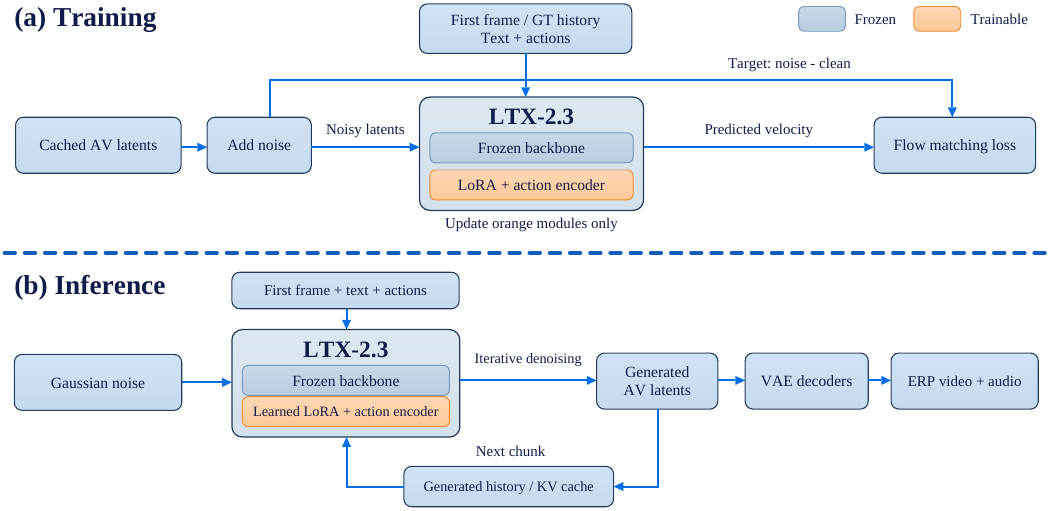}
    \caption{Training and inference for action-conditioned panoramic video and audio generation. During training, LTX-2.3 predicts audio--visual flow conditioned on the first video frame, text, and actions. During inference, autoregressive generation with cached history produces ERP video and synchronized audio.}
    \label{fig:ltx_pipeline}
\end{figure*}

\noindent\textbf{LTX-2.3 adaptation.}
We adapt pretrained LTX-2.3 for action-conditioned generation of equirectangular projection (ERP) video and synchronized audio. As illustrated in Fig.~\ref{fig:ltx_pipeline}, the video and audio VAEs, text encoder, and pretrained transformer remain frozen. We optimize only the LoRA parameters $\Delta\boldsymbol{\theta}$ and an action encoder $g_{\boldsymbol{\psi}}$. The encoder combines discrete movement and button embeddings with an MLP for continuous mouse displacements. Its outputs are temporally aligned with the video latent frame rate and injected into the timestep embeddings through adaptive layer normalization. For ERP generation, horizontal head rolling is applied to the video rotary positional embeddings (RoPE).

\noindent\textbf{Flow-matching training.}
Let $\mathbf{z}_0^{(m)}$ denote the clean latent for modality $m\in\{v,a\}$, representing video and audio, respectively. For each modality, we independently sample Gaussian noise $\boldsymbol{\epsilon}^{(m)}$ and a noise level $\sigma$:
\begin{equation}
    \mathbf{z}_{\sigma}^{(m)}
    =
    (1-\sigma)\mathbf{z}_0^{(m)}
    +\sigma\boldsymbol{\epsilon}^{(m)}.
    \label{eq:ltx_noising}
\end{equation}
Conditioned on the clean first video frame, text, and encoded actions, LTX-2.3 predicts the flow velocity $\widehat{\mathbf{v}}^{(m)}$. We optimize the masked objective
\begin{equation}
    \mathcal{L}_{\mathrm{AV}}
    =
    \mathbb{E}\left[
    \sum_{m\in\{v,a\}}\lambda_m
    \operatorname{MSE}_{M^{(m)}}\left(
    \widehat{\mathbf{v}}^{(m)},
    \boldsymbol{\epsilon}^{(m)}-\mathbf{z}_0^{(m)}
    \right)
    \right],
    \label{eq:ltx_flow_matching}
\end{equation}
where $\lambda_m$ balances the two modalities and $M^{(m)}$ denotes the corresponding binary supervision mask. The masked MSE is averaged over valid latent elements. The first video latent frame remains clean and is excluded from the loss.

Training consists of full-clip fine-tuning followed by chunk-wise autoregressive adaptation. In the latter stage, ground-truth latent history is provided through teacher forcing, while causal attention masks prevent access to future chunks. Historical latents serve only as conditioning context and are excluded from supervision. Given an initial panoramic frame, text, and action sequence, video and audio latents are generated jointly from Gaussian noise. The predicted flow is integrated from $\sigma=1$ to $\sigma=0$. During autoregressive rollout, each generated chunk is appended to the history, and its key--value states are cached for subsequent generation. The frozen video and audio decoders produce the final ERP video and synchronized audio. All parameters remain fixed during inference.

\begin{figure*}[b!]
  \includegraphics[width=0.9775\textwidth]{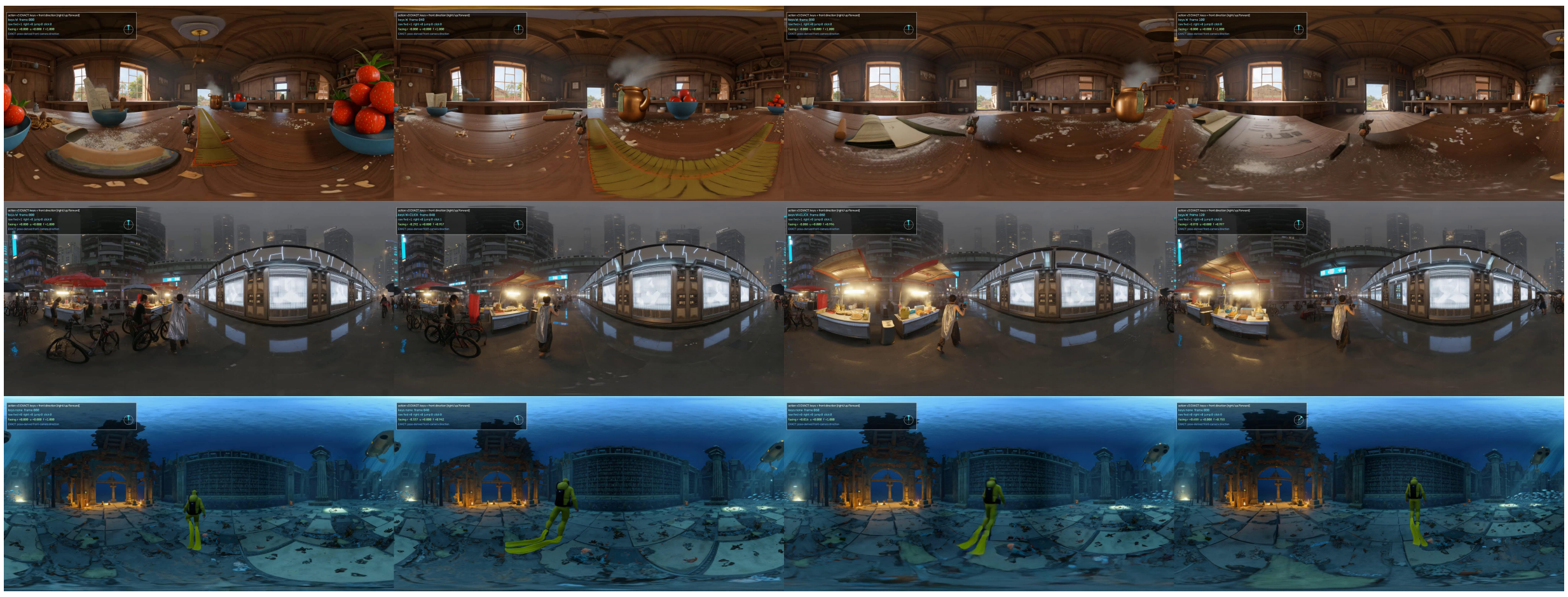}
  \caption{Panoramic generation across different scenes and protagonists.}
  \label{fig:PWM_short}
\end{figure*}
\begin{figure*}[b!]
  \includegraphics[width=0.9775\textwidth]{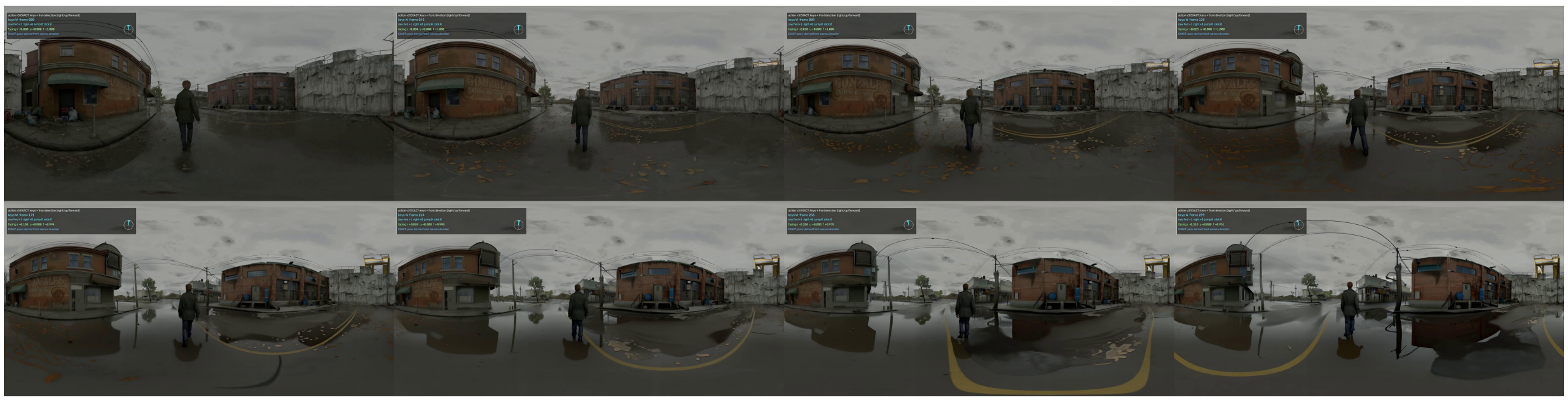}
  \caption{Long-horizon panoramic generation.}
  \label{fig:PWM_long}
\end{figure*}

\noindent\textbf{Qualitative results.}
Fig.~\ref{fig:PWM_short} presents generation results across indoor, urban, and underwater environments, demonstrating the model's applicability to diverse scenes and protagonists. Fig.~\ref{fig:PWM_long} shows an extended panoramic rollout in which scene structures remain recognizable as the viewpoint and foreground character evolve.

\section{Visual Geometry Grounded Learning}
\label{sec:VGGL}
\subsection{Cross-view alignment validation}
We evaluate cross-view alignment on 1,000 view pairs randomly sampled from the retained clips. Under the recorded alignment, the per-pair median reprojection error over co-visible points averages 0.482 pixels for multi-view recordings and 0.458 pixels for stereo recordings in Tab.~\ref{tab:cross_view_alignment}. Shifting one view forward by one frame raises these errors to 23.417 and 22.780 pixels, respectively, demonstrating sensitivity to a one-frame offset.

\begin{table}[ht]
    \centering

    \caption{Cross-view reprojection error (px) under recorded alignment and a one-frame shift.}
    \label{tab:cross_view_alignment}
    \small
    \setlength{\tabcolsep}{4pt}
    \begin{tabular}{lccc}
        \toprule
        & \multicolumn{2}{c}{Aligned} & +1-frame shift \\
        \cmidrule(lr){2-3}
        Configuration & P50 $\downarrow$ & P95 $\downarrow$ & P50$\downarrow$ \\
        \midrule
        Multi-view & 0.482 & 1.835 & 23.417 \\
        Stereo     & 0.458 & 2.193 & 22.780 \\
        \bottomrule
    \end{tabular}
\end{table}

\subsection{Stereo Reconstruction}

We initialize existing  Fast-FoundationStereo~\citep{wen2026fast} from its pretrained weights and jointly fine-tune the feature encoder, cost aggregation, and iterative update modules, while keeping batch normalization parameters frozen. Training combines public stereo datasets with GameRealm data using source-aware weighted sampling. GameRealm depth maps are converted into disparity labels using camera calibration, followed by geometric correction and validity filtering. We employ an iterative sequence-weighted \(L_1\) loss over valid pixels, AdamW optimization, and mixed-precision distributed training, with differential learning rates and cosine decay. The model is trained for 20,000 steps. We perform fine-tuning using 16 NVIDIA A800 GPUs.

Starting from the same Fast-FoundationStereo~\citep{wen2026fast} checkpoint, we fine-tune each model for 20,000 steps employing either public stereo data alone (baseline) or the same data augmented with GameRealm. We evaluate both models zero-shot on Middlebury-H and Middlebury-Q~\citep{scharstein2014high}, ETH3D~\citep{schops2017multi}, and KITTI 2015~\citep{menze2015object}. As shown in Tab.~\ref{tab:ffs_zero}, adding GameRealm reduces error on 12 of the 13 reported metrics. The largest relative reduction is 14.3\% on ETH3D BP-3, where the error drops from 0.49\% to 0.42\%.

\begin{table*}[ht]
\centering

\caption{Zero-shot stereo generalization on real-world benchmarks. Lower is better. Cell shading indicates \colorbox{bgfirst}{first}, \colorbox{bgsecond}{second}, and \colorbox{bgthird}{third} place. Baseline: public-data fine-tuning; Fast-FoundationStereo$^{*}$: public data + GameRealm.}

\label{tab:ffs_zero}
\small
\setlength{\tabcolsep}{4pt}
\renewcommand{\arraystretch}{0.9775}
\resizebox{0.9225\textwidth}{!}{%
\begin{tabular}{@{}l*{13}{c}@{}}
\toprule
\multirow{2}{*}{Method}
& \multicolumn{3}{c}{Middlebury-H}
& \multicolumn{3}{c}{Middlebury-Q}
& \multicolumn{3}{c}{ETH3D}
& \multicolumn{4}{c}{KITTI 2015} \\
\cmidrule(lr){2-4}
\cmidrule(lr){5-7}
\cmidrule(lr){8-10}
\cmidrule(lr){11-14}

& BP-1 $\downarrow$ & BP-2 $\downarrow$ & BP-3 $\downarrow$
& BP-1 $\downarrow$ & BP-2 $\downarrow$ & BP-3 $\downarrow$
& BP-1 $\downarrow$ & BP-2 $\downarrow$ & BP-3 $\downarrow$
& BP-1 $\downarrow$ & BP-2 $\downarrow$ & BP-3 $\downarrow$
& D1 $\downarrow$ \\
\midrule

Fast-FoundationStereo
& \third{4.80} & \second{2.20} & \second{1.60}
& \third{4.51} & \third{2.12} & \third{1.57}
& \third{1.22} & \third{0.62} & \third{0.50}
& \third{19.62} & \third{5.78} & \third{3.43} & \third{3.25} \\

baseline
& \second{4.76} & \third{2.22} & \third{1.63}
& \first{4.10} & \second{2.00} & \second{1.55}
& \second{1.18} & \second{0.61} & \second{0.49}
& \second{19.14} & \second{5.58} & \second{3.26} & \second{3.11} \\

Fast-FoundationStereo$^{*}$
& \first{4.66} & \first{2.03} & \first{1.42}
& \second{4.21} & \first{1.93} & \first{1.45}
& \first{1.09} & \first{0.53} & \first{0.42}
& \first{19.09} & \first{5.50} & \first{3.18} & \first{3.03} \\

\bottomrule
\end{tabular}%
}
\end{table*}

\begin{figure*}[ht]
  \includegraphics[width=0.9775\textwidth]{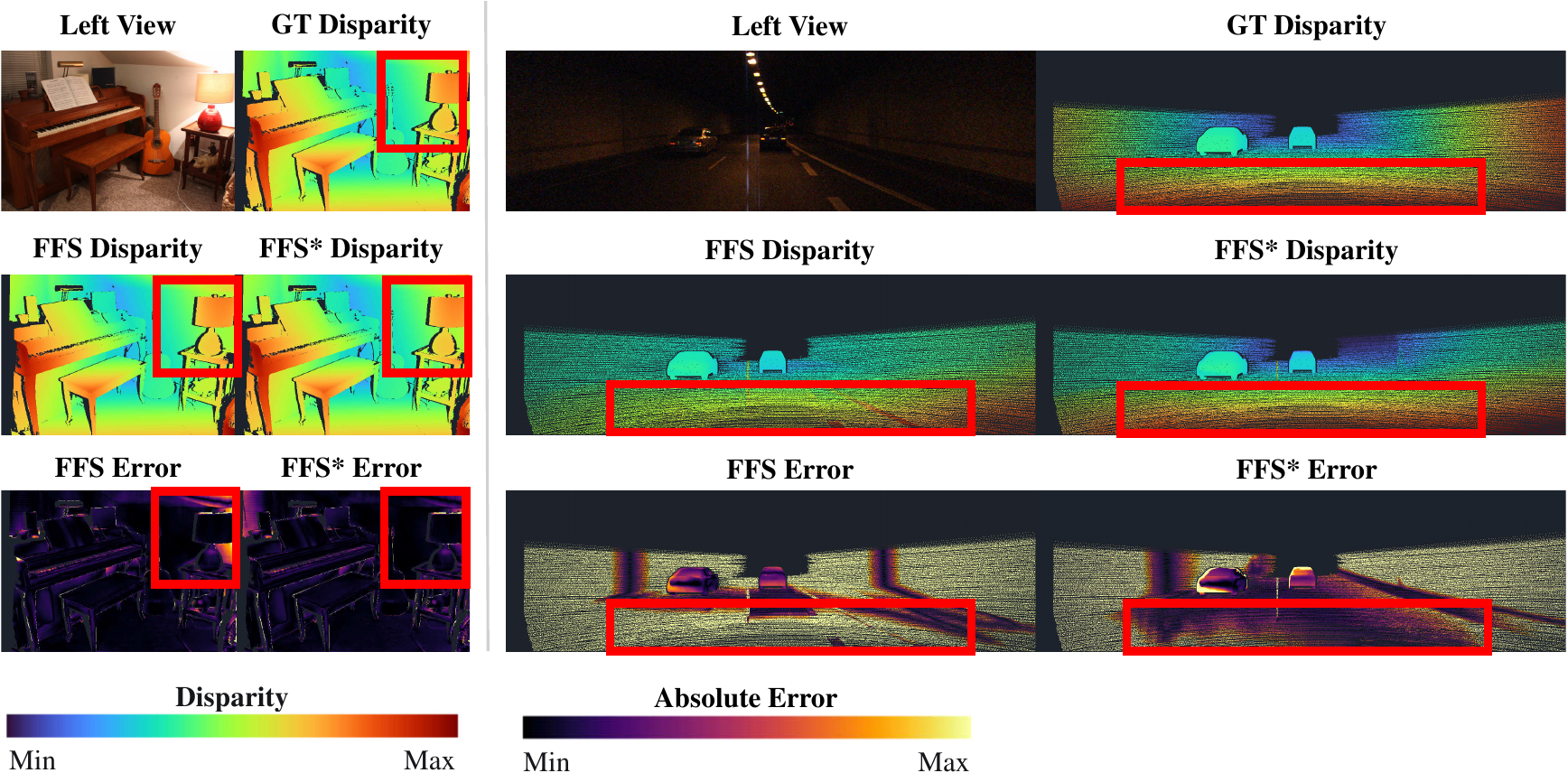}
  \caption{Stereo Finetune Comparison.}
  \label{fig:Stereo_perf}
\end{figure*}

Qualitative comparisons on indoor and outdoor scenes are presented in Fig.~\ref{fig:Stereo_perf}. In the first example, strong illumination from a desk lamp degrades the original model's reconstruction, whereas our fine-tuned model produces more accurate results. Similarly, in the second example, the original model struggles with the low-light conditions inside a tunnel, while our model achieves better reconstruction quality. These results suggest improved robustness to challenging illumination conditions, which can be attributed to the diverse visual environments provided by GameRealm.

\subsection{Multi-view Reconstruction}
To study the value of our game data for visual geometry learning, we fine-tune the released pretrained $\pi^3$ model~\citep{wang2026pi} with its original geometric losses. Our baseline uses eight datasets from the $\pi^3$ training corpus: ScanNet++~\citep{yeshwanth2023scannet++}, ARKitScenes~\citep{baruch2021arkitscenes}, MegaDepth~\citep{li2018megadepth}, CO3Dv2~\citep{reizenstein2021common}, TartanAir~\citep{wang2020tartanair}, BlendedMVG~\citep{yao2020blendedmvs}, GTA-SfM~\citep{wang2020flow}, and MatrixCity~\citep{li2023matrixcity}. We compare this baseline with a model fine-tuned on the same base datasets augmented with GameRealm. Both models share the same pretrained initialization, view-sampling strategy, and optimization settings, and are fine-tuned for 10,000 updates using AdamW with cosine learning-rate decay.

We follow the evaluation protocols in the $\pi^3$ paper~\citep{wang2026pi}, evaluating camera poses on Sintel~\citep{Butler} and RealEstate10K~\citep{zhou2018stereo}, video depth on Bonn~\citep{palazzolo2019refusion} and KITTI~\citep{geiger2013vision}, and multi-view reconstruction on 7-Scenes~\citep{shotton2013scene} and ETH3D~\citep{schops2017multi}. As shown in Table~\ref{tab:recon}, fine-tuning with GameRealm improves 11 of the 12 reported metrics over the fine-tuning baseline, including reductions of 34.7\% in Sintel ATE and 10.7\% in RPE-trans. The gains are most pronounced in camera pose estimation, while depth estimation and multi-view reconstruction show smaller improvements and mixed performance relative to the original pretrained model.

\begin{table*}[ht]
\centering
\caption{Reconstruction Quality Comparison. Cell shading denotes
\colorbox{bgfirst}{first}, \colorbox{bgsecond}{second}, and
\colorbox{bgthird}{third} place among the three models per metric. Baseline: public-data fine-tuning; $\pi^{3*}$: public data + GameRealm}
\vspace{-0.5em}
\label{tab:recon}
\footnotesize
\renewcommand{\arraystretch}{1.15}
\resizebox{\textwidth}{!}{%
\begin{tabular}{@{}l cc cc cc cc cc cc@{}}
\toprule
 & \multicolumn{4}{c}{\textbf{Camera pose}} & \multicolumn{4}{c}{\textbf{Video depth}} & \multicolumn{4}{c}{\textbf{Multi-view recon.}} \\
\cmidrule(lr){2-5}\cmidrule(lr){6-9}\cmidrule(lr){10-13}
 & \multicolumn{2}{c}{Sintel (traj.)} & \multicolumn{2}{c}{Re10K (pairwise)} & \multicolumn{2}{c}{Bonn} & \multicolumn{2}{c}{KITTI} & \multicolumn{2}{c}{7-Scenes} & \multicolumn{2}{c}{ETH3D} \\
\cmidrule(lr){2-3}\cmidrule(lr){4-5}\cmidrule(lr){6-7}\cmidrule(lr){8-9}\cmidrule(lr){10-11}\cmidrule(lr){12-13}
 & ATE $\downarrow$ & RPE-trans $\downarrow$ & AUC@5 $\uparrow$ & AUC@15 $\uparrow$
 & AbsRel $\downarrow$ & $\delta_{<1.25}$ $\uparrow$ & AbsRel $\downarrow$ & $\delta_{<1.25}$ $\uparrow$
 & Acc $\downarrow$ & Comp $\downarrow$ & Acc $\downarrow$ & Comp $\downarrow$ \\
\midrule
$\pi^3$
 & \second{0.0733} & \third{0.0390} & \second{62.66} & \second{79.93}
 & \first{0.0438} & \third{0.9753} & \third{0.0366} & \third{0.9853}
 & \first{0.0158} & \third{0.0220} & \first{0.1871} & \first{0.2104} \\
  baseline
 & \third{0.0766} & \second{0.0384} & \third{62.43} & \third{79.79}
 & \third{0.0482} & \second{0.9763} & \second{0.0363} & \first{0.9858}
 & \third{0.0166} & \second{0.0220} & \third{0.1929} & \third{0.2125} \\
$\pi^{3*}$
 & \first{0.0500} & \first{0.0343} & \first{62.96} & \first{80.18}
 & \second{0.0479} & \first{0.9778} & \first{0.0357} & \second{0.9856}
 & \second{0.0162} & \first{0.0219} & \second{0.1922} & \second{0.2112} \\
\bottomrule
\end{tabular}}
\vspace{-0.5em}
\end{table*}

\begin{figure*}[ht]
  \includegraphics[width=0.9775\textwidth]{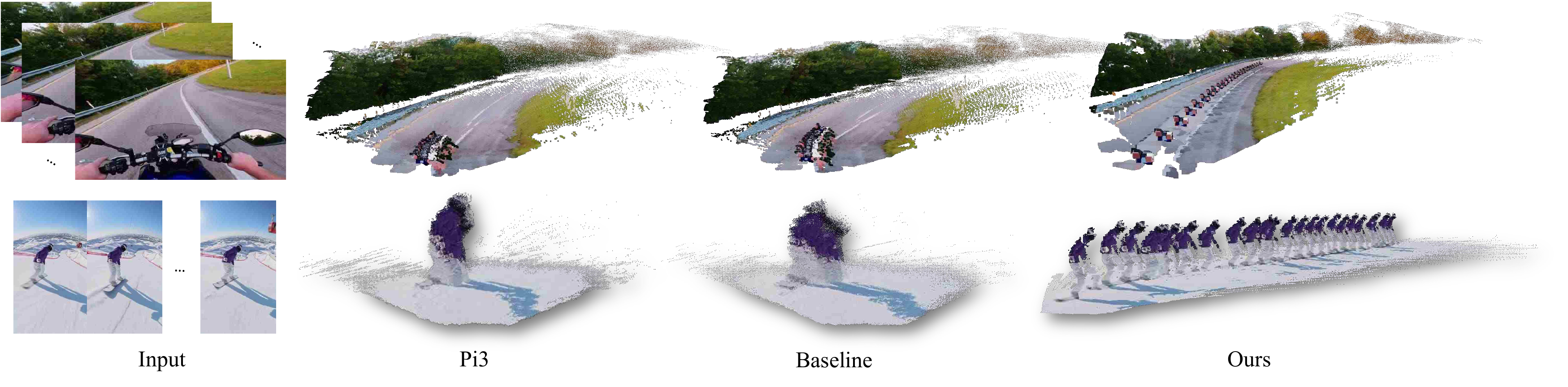}
  \caption{Reconstruction Comparison. Qualitative reconstruction comparisons on motorcycle-riding (top) and snow-sports (bottom) sequences. From left to right: sampled input frames, the original $\pi^3$, baseline fine-tuning, and fine-tuning with GameRealm (ours). Compared with the reference models, our method produces more spatially distributed reconstructions, with successive subject observations more clearly separated along the visible path.}
  \label{fig:Recon_com}
\end{figure*}

Fig.~\ref{fig:Recon_com} presents qualitative comparisons of the original $\pi^3$, baseline fine-tuning, and fine-tuning with GameRealm on motorcycle-riding and snow-sports sequences. The most apparent differences concern the spatial extent of the reconstructions and the distribution of observations across frames. In the motorcycle sequence, our method produces a more clearly delineated sequence of rider positions along the curved roadway, whereas $\pi^3$ and the fine-tuning baseline yield more spatially concentrated rider reconstructions. A similar pattern is observed in the snow-sports sequence: the two reference models produce a compact reconstruction with substantial overlap around the foreground subject, while our method reveals successive subject positions distributed over a more extended snow-covered surface. These visual differences suggest that fine-tuning with GameRealm better preserves the spatial progression of the input sequence, complementing the improvements in camera pose estimation reported in Tab.~\ref{tab:recon}.

\subsection{Point Tracking}
\label{sec:3d-tracking}

GameRealm supports both the evaluation and learning of point tracking. TrajRealm-Bench measures correspondence across diverse interactions and locomotion, while fine-tuning experiments with TAPIP3D~\citep{zhang2026tapip3d} and TrackEverything~\citep{jain2026trackeverything} assess transfer to synthetic environments and real-world video.

\noindent\textbf{Evaluation metrics and input conditions.}
Following TAPIP3D, SynthVerse, and TAPVid-3D~\citep{zhang2026tapip3d,Zhao2026SynthVerse,koppula2024tapvid}, evaluation reports average Jaccard (AJ), average percentage of points within distance thresholds (APD), and occlusion accuracy (OA). AJ jointly measures localization and visibility; APD averages localization accuracy across five thresholds; OA measures visibility classification accuracy. The benchmark and synthetic experiments report 3D and projected 2D AJ/APD together with OA; real-world evaluation reports AJ$_{3D}$, APD$_{3D}$, and OA. All scores are percentages, with higher values indicating better performance.

Benchmark methods are grouped by whether the complete tracking pipeline uses ground-truth (GT) depth and camera poses, including 3D lifting and coordinate conversion. MoSE3~\citep{cheng2026mose3}, Track4World~\citep{lu2026track4world}, and TrackEverything+$\pi^3$~\citep{jain2026trackeverything,wang2026pi} use estimated geometry. CoTracker2.1~\citep{karaev2024cotracker}, CoTracker3~\citep{karaev2025cotracker3}, DELTA~\citep{ngo2025delta}, SpatialTrackerV2~\citep{xiao2025spatialtrackerv2}, TAPIP3D, and TrackEverything+GT use GT geometry. CoTracker predicts image-space tracks and uses GT depth and poses for 3D lifting. The With GT group uses median scale normalization and depth-adaptive 3D thresholds. The Without GT group follows the MoSE3 protocol: first-frame camera coordinates, global L1 scale and translation fitting, a second median rescaling, and fixed 3D thresholds. Neither group fits rotation. Comparisons therefore use consistent scoring within each group.

\noindent\textbf{Proposed point tracking benchmark.}
TrajRealm-Bench contains 175 sequences and 104,306 camera frames across eight subsets (Table~\ref{tab:gamerealm-tracking-splits}). Interaction evaluates point identity through contact, articulation, and occlusion: General covers everyday human, animal, and object interactions; Hard covers rapid motion and large pose changes; Embodied covers robotic manipulation; and Plant covers deforming leaves and branches. Locomotion evaluates movement through environments: General covers individual movement and navigation; Transportation covers driving and riding; Group covers intersecting trajectories and mutual occlusion; and Long-horizon covers sustained movement over more than 800 temporal frames.

Each sequence belongs to one subset, with Long-horizon sequences assigned exclusively to that subset. Recording-group splits keep synchronized views and adjacent windows together. Evaluation retains full sequences, with up to 256 unique annotated queries per camera input, initialized at their first eligible visible frame. The 191 camera inputs provide 42,173 queries. AJ and OA use a fixed visibility threshold of 0.5 on this benchmark. Scores are averaged across views within each source sequence and then across sequences within each subset; family and overall averages weight their four and eight subsets equally. CoTracker2.1, DELTA, and TAPIP3D use 16-frame windows with stride 8; SpatialTrackerV2 uses length/stride 200/196 offline and 20/14 online; TrackEverything uses 16/16 blocks with persistent scene memory; CoTracker3-offline processes the full sequence jointly. TrackEverything+$\pi^3$ estimates geometry in 60-frame chunks with overlap 30.

\begin{table}[htbp] 
    \centering
    \caption{\textbf{Proposed benchmark statistics.}}
    \label{tab:gamerealm-tracking-splits}
    \footnotesize 
    \setlength{\tabcolsep}{5pt}
    \renewcommand{\arraystretch}{1.15} 
    
    \begin{tabular}{@{}lrr@{\hspace{2em}}lrr@{}}
        \toprule
        \multicolumn{3}{c}{\textbf{Interaction}} & \multicolumn{3}{c}{\textbf{Locomotion}} \\
        \midrule
        Category & Seq. & Frames & Category & Seq. & Frames \\
        \midrule
        General  & 40 & 8,495 & General        & 30 & 7,066 \\
        Hard     & 15 & 3,904 & Long-horizon   & 20 & 61,718 \\
        Embodied & 15 & 6,123 & Transportation & 30 & 11,492 \\
        Plant    & 15 & 3,690 & Group          & 10 & 1,818 \\
        \midrule
        Subtotal & 85 & 22,212 & Subtotal & 90 & 82,094 \\ 
        \midrule
        \multicolumn{6}{c}{\textbf{Overall Total: 175 Sequences, 104,306 Frames}} \\
        \bottomrule
    \end{tabular}
\end{table}

\begin{table*}[!t]
\centering
\caption{\textbf{Proposed TrajRealm-Bench results.} $^{*}$: fine-tuned; $^{\dagger}$: dense. Dark, medium, and light yellow mark the top three distinct displayed scores within each GT group and metric; ties share a rank. GT groups include lifting and coordinate conversion; $\pi^3$: estimated geometry. Track4World ($^{\ddagger}$) scores and averages exclude sequences with non-finite (NaN/Inf) predictions. Its successful/total sequence counts are Interaction---General 40/40, Hard 14/15, Embodied 14/15, Plant 15/15; Locomotion---General 29/30, Long-horizon 2/20, Transportation 14/30, Group 9/10.}
\label{tab:gamerealm-benchmark}
\setlength{\tabcolsep}{3pt}
\newcommand{\blankscore}{--}
\renewcommand{\arraystretch}{1.05}
\newsavebox{\gamerealmbenchmarkpanel}
\sbox{\gamerealmbenchmarkpanel}{%
\begin{tabular}{l|ccccc|ccccc|ccccc|ccccc}
\toprule
\multirow{2}{*}{\shortstack[l]{\textbf{\textit{Interaction}}\\Methods}} & \multicolumn{5}{c|}{General} & \multicolumn{5}{c|}{Hard} & \multicolumn{5}{c|}{Embodied} & \multicolumn{5}{c}{Plant} \\
 & $\mathrm{AJ}_{3D}\uparrow$ & $\mathrm{APD}_{3D}\uparrow$ & $\mathrm{AJ}_{2D}\uparrow$ & $\mathrm{APD}_{2D}\uparrow$ & $\mathrm{OA}\uparrow$ & $\mathrm{AJ}_{3D}\uparrow$ & $\mathrm{APD}_{3D}\uparrow$ & $\mathrm{AJ}_{2D}\uparrow$ & $\mathrm{APD}_{2D}\uparrow$ & $\mathrm{OA}\uparrow$ & $\mathrm{AJ}_{3D}\uparrow$ & $\mathrm{APD}_{3D}\uparrow$ & $\mathrm{AJ}_{2D}\uparrow$ & $\mathrm{APD}_{2D}\uparrow$ & $\mathrm{OA}\uparrow$ & $\mathrm{AJ}_{3D}\uparrow$ & $\mathrm{APD}_{3D}\uparrow$ & $\mathrm{AJ}_{2D}\uparrow$ & $\mathrm{APD}_{2D}\uparrow$ & $\mathrm{OA}\uparrow$ \\
\midrule
\multicolumn{21}{l}{\textbf{\textit{Without GT depth and camera pose}}} \\
Track4World$^{\dagger}$ & \cellcolor{bgthird}16.2 & 30.8 & \cellcolor{bgfirst}34.0 & \cellcolor{bgfirst}56.0 & \cellcolor{bgsecond}77.7 & 3.9\rlap{$^{\ddagger}$} & 12.6\rlap{$^{\ddagger}$} & \cellcolor{bgfirst}20.6\rlap{$^{\ddagger}$} & \cellcolor{bgfirst}48.2\rlap{$^{\ddagger}$} & \cellcolor{bgfirst}69.7\rlap{$^{\ddagger}$} & 52.7\rlap{$^{\ddagger}$} & 63.1\rlap{$^{\ddagger}$} & \cellcolor{bgfirst}77.4\rlap{$^{\ddagger}$} & \cellcolor{bgfirst}88.4\rlap{$^{\ddagger}$} & \cellcolor{bgsecond}91.3\rlap{$^{\ddagger}$} & 19.7 & 35.8 & \cellcolor{bgfirst}44.2 & \cellcolor{bgfirst}78.6 & 60.8 \\
MoSE3$^{\dagger}$ & \cellcolor{bgfirst}22.8 & \cellcolor{bgfirst}41.5 & \cellcolor{bgthird}22.8 & \cellcolor{bgthird}41.2 & \cellcolor{bgthird}76.2 & \cellcolor{bgfirst}10.8 & \cellcolor{bgfirst}27.3 & \cellcolor{bgsecond}13.1 & \cellcolor{bgsecond}35.6 & \cellcolor{bgthird}65.2 & \cellcolor{bgthird}55.2 & \cellcolor{bgthird}65.8 & \cellcolor{bgsecond}63.1 & \cellcolor{bgthird}76.0 & \cellcolor{bgfirst}93.0 & \cellcolor{bgfirst}22.5 & \cellcolor{bgfirst}40.5 & \cellcolor{bgsecond}38.8 & \cellcolor{bgsecond}70.0 & \cellcolor{bgthird}61.8 \\
{TrackEverything +$\pi^3$}$^{\dagger}$ & \cellcolor{bgthird}16.2 & \cellcolor{bgthird}35.6 & 15.3 & 30.9 & 64.1 & \cellcolor{bgthird}6.6 & \cellcolor{bgthird}20.1 & 4.6 & 15.7 & 46.9 & \cellcolor{bgfirst}56.6 & \cellcolor{bgfirst}70.3 & \cellcolor{bgthird}62.0 & \cellcolor{bgsecond}77.0 & \cellcolor{bgthird}89.6 & \cellcolor{bgsecond}21.8 & \cellcolor{bgsecond}39.9 & 32.0 & 58.4 & \cellcolor{bgsecond}63.7 \\
\specialrule{\lightrulewidth}{0.5pt}{0.5pt}
TrackEverything +$\pi^3$$^{*\dagger}$ & \cellcolor{bgsecond}21.7 & \cellcolor{bgsecond}40.9 & \cellcolor{bgsecond}24.2 & \cellcolor{bgsecond}44.1 & \cellcolor{bgfirst}78.2 & \cellcolor{bgsecond}10.1 & \cellcolor{bgsecond}26.5 & \cellcolor{bgthird}12.7 & \cellcolor{bgthird}32.4 & \cellcolor{bgsecond}68.3 & \cellcolor{bgsecond}55.9 & \cellcolor{bgsecond}69.2 & 59.9 & 74.5 & \cellcolor{bgthird}89.6 & \cellcolor{bgthird}21.3 & \cellcolor{bgthird}39.1 & \cellcolor{bgthird}34.6 & \cellcolor{bgthird}63.4 & \cellcolor{bgfirst}65.4 \\
\addlinespace[2pt]
\multicolumn{21}{l}{\textbf{\textit{With GT depth and camera pose}}} \\
CoTracker2.1 & 14.6 & 27.5 & 22.0 & 42.5 & 68.7 & 8.7 & 22.8 & 12.7 & 34.4 & 62.5 & 50.2 & 61.6 & 68.7 & 82.6 & 84.3 & 13.2 & 25.0 & 37.5 & 73.4 & 58.4 \\
CoTracker3 & \cellcolor{bgsecond}19.6 & \cellcolor{bgthird}35.3 & \cellcolor{bgthird}29.0 & \cellcolor{bgthird}53.4 & \cellcolor{bgthird}75.9 & \cellcolor{bgthird}12.8 & \cellcolor{bgthird}32.4 & \cellcolor{bgthird}16.3 & \cellcolor{bgthird}48.1 & \cellcolor{bgthird}72.8 & 57.3 & 68.9 & 74.3 & 86.9 & 88.3 & 13.7 & 24.7 & 43.5 & \cellcolor{bgthird}76.9 & 59.6 \\
DELTA$^{\dagger}$ & 14.7 & 25.6 & 27.9 & 47.5 & 72.6 & 8.3 & 21.3 & 15.8 & 41.3 & 58.2 & 43.8 & 53.9 & 76.1 & 86.7 & 91.6 & 21.2 & 34.9 & \cellcolor{bgthird}45.0 & 73.7 & 70.2 \\
SpatialTrackerV2-offline & 14.8 & 24.3 & 23.9 & 46.1 & 69.3 & 5.5 & 16.7 & 11.3 & 34.2 & 55.6 & 51.6 & 63.9 & 63.5 & 81.8 & 83.8 & 4.0 & 8.9 & 33.0 & 60.8 & 59.6 \\
SpatialTrackerV2-online & 11.7 & 27.5 & 20.1 & 48.4 & 52.0 & 6.5 & 24.5 & 11.9 & 44.1 & 37.9 & 42.3 & 58.6 & 60.1 & 80.9 & 82.1 & 6.9 & 13.8 & 36.5 & 68.1 & 59.1 \\
TrackEverything +GT$^{\dagger}$ & 13.2 & 23.8 & 19.1 & 36.1 & 68.4 & 1.6 & 4.1 & 3.8 & 11.2 & 51.6 & 67.4 & 79.2 & \cellcolor{bgthird}76.4 & 89.1 & \cellcolor{bgfirst}93.0 & 30.4 & 50.9 & 41.6 & 69.5 & \cellcolor{bgthird}70.8 \\
TAPIP3D-camera & 14.6 & 30.7 & 19.5 & 42.4 & 53.3 & 4.1 & 13.3 & 7.4 & 24.6 & 36.8 & \cellcolor{bgthird}70.1 & 86.2 & 73.9 & 89.4 & 90.6 & \cellcolor{bgthird}32.4 & 57.5 & 40.3 & 71.5 & 58.7 \\
TAPIP3D-world & \cellcolor{bgthird}15.7 & 32.0 & 20.1 & 42.6 & 55.4 & 4.4 & 15.1 & 7.6 & 25.8 & 36.2 & 70.0 & \cellcolor{bgsecond}86.4 & 74.3 & \cellcolor{bgthird}89.8 & 90.7 & 32.1 & \cellcolor{bgthird}58.0 & 40.2 & 72.0 & 58.3 \\
\specialrule{\lightrulewidth}{0.5pt}{0.5pt}
TAPIP3D-camera$^{*}$ & \cellcolor{bgfirst}32.0 & \cellcolor{bgfirst}47.0 & \cellcolor{bgfirst}38.7 & \cellcolor{bgfirst}59.4 & \cellcolor{bgfirst}82.5 & \cellcolor{bgfirst}22.6 & \cellcolor{bgsecond}41.6 & \cellcolor{bgfirst}27.2 & \cellcolor{bgsecond}54.7 & \cellcolor{bgsecond}76.7 & \cellcolor{bgsecond}74.5 & \cellcolor{bgthird}86.3 & \cellcolor{bgsecond}77.4 & \cellcolor{bgsecond}90.0 & \cellcolor{bgthird}91.9 & \cellcolor{bgfirst}44.2 & \cellcolor{bgfirst}66.6 & \cellcolor{bgfirst}49.8 & \cellcolor{bgfirst}79.5 & \cellcolor{bgsecond}72.7 \\
TAPIP3D-world$^{*}$ & \cellcolor{bgfirst}32.0 & \cellcolor{bgsecond}45.0 & \cellcolor{bgsecond}37.4 & \cellcolor{bgsecond}55.1 & \cellcolor{bgsecond}80.9 & \cellcolor{bgsecond}21.4 & \cellcolor{bgfirst}41.7 & \cellcolor{bgsecond}26.7 & \cellcolor{bgfirst}55.4 & \cellcolor{bgfirst}77.0 & \cellcolor{bgfirst}75.0 & \cellcolor{bgfirst}86.6 & \cellcolor{bgfirst}78.2 & \cellcolor{bgfirst}90.4 & \cellcolor{bgsecond}92.4 & \cellcolor{bgsecond}43.8 & \cellcolor{bgsecond}64.9 & \cellcolor{bgsecond}49.7 & \cellcolor{bgsecond}78.9 & \cellcolor{bgfirst}73.2 \\
\end{tabular}}
\resizebox{\textwidth}{!}{%
\begin{minipage}{\wd\gamerealmbenchmarkpanel}
\setlength{\parindent}{0pt}
\usebox{\gamerealmbenchmarkpanel}
\par\vspace{1pt}
\begin{tabular*}{\linewidth}{l|@{\hspace{\tabcolsep}\extracolsep{\fill}}ccccc|ccccc|ccccc|ccccc}
\midrule
\multirow{2}{*}{\shortstack[l]{\textbf{\textit{Locomotion}}\\Methods}} & \multicolumn{5}{c|}{General} & \multicolumn{5}{c|}{Long-horizon} & \multicolumn{5}{c|}{Transportation} & \multicolumn{5}{c}{Group} \\
 & $\mathrm{AJ}_{3D}\uparrow$ & $\mathrm{APD}_{3D}\uparrow$ & $\mathrm{AJ}_{2D}\uparrow$ & $\mathrm{APD}_{2D}\uparrow$ & $\mathrm{OA}\uparrow$ & $\mathrm{AJ}_{3D}\uparrow$ & $\mathrm{APD}_{3D}\uparrow$ & $\mathrm{AJ}_{2D}\uparrow$ & $\mathrm{APD}_{2D}\uparrow$ & $\mathrm{OA}\uparrow$ & $\mathrm{AJ}_{3D}\uparrow$ & $\mathrm{APD}_{3D}\uparrow$ & $\mathrm{AJ}_{2D}\uparrow$ & $\mathrm{APD}_{2D}\uparrow$ & $\mathrm{OA}\uparrow$ & $\mathrm{AJ}_{3D}\uparrow$ & $\mathrm{APD}_{3D}\uparrow$ & $\mathrm{AJ}_{2D}\uparrow$ & $\mathrm{APD}_{2D}\uparrow$ & $\mathrm{OA}\uparrow$ \\
\midrule
\multicolumn{21}{l}{\textbf{\textit{Without GT depth and camera pose}}} \\
Track4World$^{\dagger}$ & 12.0\rlap{$^{\ddagger}$} & 16.5\rlap{$^{\ddagger}$} & \cellcolor{bgfirst}41.4\rlap{$^{\ddagger}$} & \cellcolor{bgfirst}61.0\rlap{$^{\ddagger}$} & \cellcolor{bgfirst}79.9\rlap{$^{\ddagger}$} & \cellcolor{bgthird}0.5\rlap{$^{\ddagger}$} & \cellcolor{bgthird}1.9\rlap{$^{\ddagger}$} & \cellcolor{bgfirst}14.1\rlap{$^{\ddagger}$} & \cellcolor{bgfirst}30.1\rlap{$^{\ddagger}$} & \cellcolor{bgfirst}74.9\rlap{$^{\ddagger}$} & 1.0\rlap{$^{\ddagger}$} & 2.5\rlap{$^{\ddagger}$} & \cellcolor{bgfirst}30.9\rlap{$^{\ddagger}$} & \cellcolor{bgfirst}55.9\rlap{$^{\ddagger}$} & \cellcolor{bgfirst}70.6\rlap{$^{\ddagger}$} & 11.5\rlap{$^{\ddagger}$} & 20.4\rlap{$^{\ddagger}$} & \cellcolor{bgfirst}40.9\rlap{$^{\ddagger}$} & \cellcolor{bgfirst}62.8\rlap{$^{\ddagger}$} & \cellcolor{bgfirst}76.6\rlap{$^{\ddagger}$} \\
MoSE3$^{\dagger}$ & \cellcolor{bgthird}17.4 & \cellcolor{bgthird}24.0 & \cellcolor{bgthird}26.6 & \cellcolor{bgthird}41.6 & \cellcolor{bgsecond}79.6 & \cellcolor{bgfirst}6.8 & \cellcolor{bgfirst}15.4 & \cellcolor{bgsecond}10.6 & \cellcolor{bgsecond}28.3 & \cellcolor{bgthird}65.6 & \cellcolor{bgsecond}4.5 & \cellcolor{bgsecond}9.7 & \cellcolor{bgsecond}15.9 & \cellcolor{bgsecond}33.2 & \cellcolor{bgsecond}66.9 & \cellcolor{bgthird}15.3 & \cellcolor{bgthird}27.3 & \cellcolor{bgthird}24.0 & 40.7 & \cellcolor{bgsecond}74.4 \\
{TrackEverything +$\pi^3$}$^{\dagger}$ & \cellcolor{bgsecond}19.0 & \cellcolor{bgsecond}25.1 & 25.4 & 34.0 & 72.4 & 0.4 & 1.7 & 1.6 & 3.6 & 61.7 & \cellcolor{bgthird}2.6 & \cellcolor{bgthird}5.3 & 5.6 & 11.1 & 59.0 & \cellcolor{bgsecond}15.7 & \cellcolor{bgsecond}30.9 & 23.6 & \cellcolor{bgthird}46.4 & 66.6 \\
\specialrule{\lightrulewidth}{0.5pt}{0.5pt}
TrackEverything +$\pi^3$$^{*\dagger}$ & \cellcolor{bgfirst}21.3 & \cellcolor{bgfirst}29.1 & \cellcolor{bgsecond}30.4 & \cellcolor{bgsecond}44.0 & \cellcolor{bgthird}79.0 & \cellcolor{bgsecond}1.7 & \cellcolor{bgsecond}4.8 & \cellcolor{bgthird}5.4 & \cellcolor{bgthird}14.5 & \cellcolor{bgsecond}66.9 & \cellcolor{bgfirst}5.1 & \cellcolor{bgfirst}10.2 & \cellcolor{bgthird}14.4 & \cellcolor{bgthird}29.2 & \cellcolor{bgthird}65.2 & \cellcolor{bgfirst}17.3 & \cellcolor{bgfirst}32.2 & \cellcolor{bgsecond}29.3 & \cellcolor{bgsecond}53.6 & \cellcolor{bgthird}70.9 \\
\addlinespace[2pt]
\multicolumn{21}{l}{\textbf{\textit{With GT depth and camera pose}}} \\
CoTracker2.1 & 21.0 & 33.4 & 32.3 & 51.4 & 75.1 & 6.9 & 14.4 & 13.1 & 28.6 & 62.6 & 17.1 & 32.6 & 31.1 & 58.3 & 69.6 & 9.1 & 16.8 & 34.4 & 60.5 & 72.0 \\
CoTracker3 & 22.3 & 35.1 & \cellcolor{bgthird}35.7 & \cellcolor{bgthird}56.9 & \cellcolor{bgthird}78.0 & \cellcolor{bgthird}12.5 & \cellcolor{bgthird}29.3 & \cellcolor{bgthird}19.8 & \cellcolor{bgsecond}48.0 & \cellcolor{bgthird}69.5 & 21.0 & 39.4 & \cellcolor{bgthird}35.8 & \cellcolor{bgthird}65.4 & \cellcolor{bgthird}74.4 & 10.0 & 17.3 & 38.4 & 64.1 & 75.3 \\
DELTA$^{\dagger}$ & 10.2 & 16.1 & 30.3 & 43.5 & 74.0 & 3.3 & 7.5 & 13.3 & 29.0 & 61.8 & 11.7 & 23.5 & \cellcolor{bgthird}35.8 & 63.9 & 70.1 & 12.2 & 22.5 & \cellcolor{bgthird}40.2 & 66.0 & \cellcolor{bgthird}76.4 \\
SpatialTrackerV2-offline & 16.7 & 24.1 & 30.8 & 49.6 & 74.6 & 3.6 & 9.1 & 11.7 & 35.4 & 47.5 & 5.1 & 9.8 & 19.9 & 45.2 & 60.3 & 2.5 & 3.6 & 24.6 & 47.0 & 68.2 \\
SpatialTrackerV2-online & 11.2 & 21.0 & 24.8 & 48.7 & 60.2 & 5.4 & 17.5 & 12.6 & 41.3 & 38.6 & 9.9 & 25.0 & 22.2 & 55.0 & 46.6 & 5.6 & 15.2 & 28.0 & 57.3 & 61.0 \\
TrackEverything +GT$^{\dagger}$ & 28.0 & 32.1 & 32.5 & 38.2 & 73.0 & 0.2 & 0.4 & 1.7 & 4.0 & 62.5 & 1.4 & 2.8 & 5.9 & 11.6 & 59.3 & 17.7 & 28.6 & 31.2 & 56.4 & 70.4 \\
TAPIP3D-camera & 23.1 & 35.0 & 30.2 & 47.6 & 65.9 & 8.3 & 22.6 & 13.7 & 40.0 & 46.3 & \cellcolor{bgthird}24.9 & \cellcolor{bgthird}50.5 & 31.4 & \cellcolor{bgthird}65.4 & 56.9 & 22.9 & 39.5 & 35.3 & 65.7 & 67.2 \\
TAPIP3D-world & \cellcolor{bgthird}30.1 & \cellcolor{bgthird}43.4 & 34.9 & 51.9 & 67.5 & 7.4 & 20.2 & 12.4 & 35.4 & 49.3 & 23.2 & 47.4 & 29.1 & 60.9 & 56.2 & \cellcolor{bgthird}23.6 & \cellcolor{bgthird}43.0 & 36.1 & \cellcolor{bgthird}67.0 & 67.2 \\
\specialrule{\lightrulewidth}{0.5pt}{0.5pt}
TAPIP3D-camera$^{*}$ & \cellcolor{bgsecond}38.2 & \cellcolor{bgsecond}53.5 & \cellcolor{bgsecond}46.9 & \cellcolor{bgsecond}66.5 & \cellcolor{bgfirst}84.8 & \cellcolor{bgfirst}27.2 & \cellcolor{bgfirst}42.0 & \cellcolor{bgfirst}33.5 & \cellcolor{bgfirst}58.5 & \cellcolor{bgfirst}80.4 & \cellcolor{bgfirst}44.1 & \cellcolor{bgfirst}63.8 & \cellcolor{bgfirst}51.5 & \cellcolor{bgfirst}75.7 & \cellcolor{bgfirst}84.9 & \cellcolor{bgsecond}39.1 & \cellcolor{bgsecond}54.9 & \cellcolor{bgsecond}50.1 & \cellcolor{bgsecond}76.2 & \cellcolor{bgsecond}81.4 \\
TAPIP3D-world$^{*}$ & \cellcolor{bgfirst}42.6 & \cellcolor{bgfirst}56.9 & \cellcolor{bgfirst}49.1 & \cellcolor{bgfirst}66.6 & \cellcolor{bgsecond}84.5 & \cellcolor{bgsecond}20.3 & \cellcolor{bgsecond}30.7 & \cellcolor{bgsecond}26.3 & \cellcolor{bgthird}45.5 & \cellcolor{bgsecond}76.8 & \cellcolor{bgsecond}38.1 & \cellcolor{bgsecond}55.4 & \cellcolor{bgsecond}46.1 & \cellcolor{bgsecond}68.0 & \cellcolor{bgsecond}82.5 & \cellcolor{bgfirst}43.1 & \cellcolor{bgfirst}63.3 & \cellcolor{bgfirst}50.7 & \cellcolor{bgfirst}76.7 & \cellcolor{bgfirst}81.9 \\
\end{tabular*}
\par\vspace{1pt}
\begin{tabular*}{\linewidth}{l|@{\hspace{\tabcolsep}\extracolsep{\fill}}ccccc|ccccc|ccccc}
\midrule
\multirow{2}{*}{\shortstack[l]{\textbf{\textit{Averages}}\\Methods}} & \multicolumn{5}{c|}{Interaction Avg.} & \multicolumn{5}{c|}{Locomotion Avg.} & \multicolumn{5}{c}{Overall Avg.} \\
 & $\mathrm{AJ}_{3D}\uparrow$ & $\mathrm{APD}_{3D}\uparrow$ & $\mathrm{AJ}_{2D}\uparrow$ & $\mathrm{APD}_{2D}\uparrow$ & $\mathrm{OA}\uparrow$ & $\mathrm{AJ}_{3D}\uparrow$ & $\mathrm{APD}_{3D}\uparrow$ & $\mathrm{AJ}_{2D}\uparrow$ & $\mathrm{APD}_{2D}\uparrow$ & $\mathrm{OA}\uparrow$ & $\mathrm{AJ}_{3D}\uparrow$ & $\mathrm{APD}_{3D}\uparrow$ & $\mathrm{AJ}_{2D}\uparrow$ & $\mathrm{APD}_{2D}\uparrow$ & $\mathrm{OA}\uparrow$ \\
\midrule
\multicolumn{16}{l}{\textbf{\textit{Without GT depth and camera pose}}} \\
Track4World$^{\dagger}$ & 23.1\rlap{$^{\ddagger}$} & 35.6\rlap{$^{\ddagger}$} & \cellcolor{bgfirst}44.1\rlap{$^{\ddagger}$} & \cellcolor{bgfirst}67.8\rlap{$^{\ddagger}$} & \cellcolor{bgsecond}74.9\rlap{$^{\ddagger}$} & 6.3\rlap{$^{\ddagger}$} & \cellcolor{bgthird}10.3\rlap{$^{\ddagger}$} & \cellcolor{bgfirst}31.8\rlap{$^{\ddagger}$} & \cellcolor{bgfirst}52.4\rlap{$^{\ddagger}$} & \cellcolor{bgfirst}75.5\rlap{$^{\ddagger}$} & 14.7\rlap{$^{\ddagger}$} & 22.9\rlap{$^{\ddagger}$} & \cellcolor{bgfirst}37.9\rlap{$^{\ddagger}$} & \cellcolor{bgfirst}60.1\rlap{$^{\ddagger}$} & \cellcolor{bgfirst}75.2\rlap{$^{\ddagger}$} \\
MoSE3$^{\dagger}$ & \cellcolor{bgfirst}27.8 & \cellcolor{bgsecond}43.8 & \cellcolor{bgsecond}34.4 & \cellcolor{bgsecond}55.7 & \cellcolor{bgthird}74.1 & \cellcolor{bgsecond}11.0 & \cellcolor{bgfirst}19.1 & \cellcolor{bgthird}19.3 & \cellcolor{bgsecond}35.9 & \cellcolor{bgsecond}71.6 & \cellcolor{bgfirst}19.4 & \cellcolor{bgsecond}31.4 & \cellcolor{bgsecond}26.9 & \cellcolor{bgsecond}45.8 & \cellcolor{bgthird}72.8 \\
{TrackEverything +$\pi^3$}$^{\dagger}$ & \cellcolor{bgthird}25.3 & \cellcolor{bgthird}41.5 & 28.5 & 45.5 & 66.1 & \cellcolor{bgthird}9.4 & \cellcolor{bgsecond}15.7 & 14.0 & 23.8 & 64.9 & \cellcolor{bgthird}17.4 & \cellcolor{bgthird}28.6 & 21.3 & 34.6 & 65.5 \\
\specialrule{\lightrulewidth}{0.5pt}{0.5pt}
TrackEverything +$\pi^3$$^{*\dagger}$ & \cellcolor{bgsecond}27.3 & \cellcolor{bgfirst}43.9 & \cellcolor{bgthird}32.9 & \cellcolor{bgthird}53.6 & \cellcolor{bgfirst}75.4 & \cellcolor{bgfirst}11.4 & \cellcolor{bgfirst}19.1 & \cellcolor{bgsecond}19.9 & \cellcolor{bgthird}35.3 & \cellcolor{bgthird}70.5 & \cellcolor{bgsecond}19.3 & \cellcolor{bgfirst}31.5 & \cellcolor{bgthird}26.4 & \cellcolor{bgthird}44.5 & \cellcolor{bgsecond}72.9 \\
\addlinespace[2pt]
\multicolumn{16}{l}{\textbf{\textit{With GT depth and camera pose}}} \\
CoTracker2.1 & 21.7 & 34.2 & 35.2 & 58.2 & 68.5 & 13.5 & 24.3 & 27.7 & 49.7 & 69.8 & 17.6 & 29.3 & 31.5 & 54.0 & 69.1 \\
CoTracker3 & 25.9 & 40.3 & 40.7 & \cellcolor{bgthird}66.3 & \cellcolor{bgthird}74.2 & 16.4 & 30.3 & \cellcolor{bgthird}32.4 & \cellcolor{bgthird}58.6 & \cellcolor{bgthird}74.3 & 21.1 & 35.3 & \cellcolor{bgthird}36.6 & \cellcolor{bgthird}62.5 & \cellcolor{bgthird}74.2 \\
DELTA$^{\dagger}$ & 22.0 & 33.9 & \cellcolor{bgthird}41.2 & 62.3 & 73.1 & 9.3 & 17.4 & 29.9 & 50.6 & 70.6 & 15.7 & 25.7 & 35.5 & 56.4 & 71.9 \\
SpatialTrackerV2-offline & 19.0 & 28.4 & 32.9 & 55.7 & 67.0 & 7.0 & 11.6 & 21.7 & 44.3 & 62.7 & 13.0 & 20.0 & 27.3 & 50.0 & 64.9 \\
SpatialTrackerV2-online & 16.9 & 31.1 & 32.1 & 60.4 & 57.7 & 8.0 & 19.7 & 21.9 & 50.6 & 51.6 & 12.4 & 25.4 & 27.0 & 55.5 & 54.7 \\
TrackEverything +GT$^{\dagger}$ & 28.1 & 39.5 & 35.2 & 51.5 & 70.9 & 11.8 & 16.0 & 17.8 & 27.6 & 66.3 & 20.0 & 27.7 & 26.5 & 39.5 & 68.6 \\
TAPIP3D-camera & 30.3 & 46.9 & 35.3 & 57.0 & 59.9 & 19.8 & 36.9 & 27.6 & 54.7 & 59.1 & 25.1 & 41.9 & 31.5 & 55.8 & 59.5 \\
TAPIP3D-world & \cellcolor{bgthird}30.5 & \cellcolor{bgthird}47.9 & 35.5 & 57.6 & 60.1 & \cellcolor{bgthird}21.1 & \cellcolor{bgthird}38.5 & 28.1 & 53.8 & 60.1 & \cellcolor{bgthird}25.8 & \cellcolor{bgthird}43.2 & 31.8 & 55.7 & 60.1 \\
\specialrule{\lightrulewidth}{0.5pt}{0.5pt}
TAPIP3D-camera$^{*}$ & \cellcolor{bgfirst}43.3 & \cellcolor{bgfirst}60.3 & \cellcolor{bgfirst}48.3 & \cellcolor{bgfirst}70.9 & \cellcolor{bgfirst}81.0 & \cellcolor{bgfirst}37.2 & \cellcolor{bgfirst}53.6 & \cellcolor{bgfirst}45.5 & \cellcolor{bgfirst}69.2 & \cellcolor{bgfirst}82.9 & \cellcolor{bgfirst}40.2 & \cellcolor{bgfirst}57.0 & \cellcolor{bgfirst}46.9 & \cellcolor{bgfirst}70.1 & \cellcolor{bgfirst}81.9 \\
TAPIP3D-world$^{*}$ & \cellcolor{bgsecond}43.0 & \cellcolor{bgsecond}59.5 & \cellcolor{bgsecond}48.0 & \cellcolor{bgsecond}69.9 & \cellcolor{bgsecond}80.9 & \cellcolor{bgsecond}36.0 & \cellcolor{bgsecond}51.6 & \cellcolor{bgsecond}43.0 & \cellcolor{bgsecond}64.2 & \cellcolor{bgsecond}81.4 & \cellcolor{bgsecond}39.5 & \cellcolor{bgsecond}55.6 & \cellcolor{bgsecond}45.5 & \cellcolor{bgsecond}67.1 & \cellcolor{bgsecond}81.2 \\
\bottomrule
\end{tabular*}
\end{minipage}}
\end{table*}

\noindent\textbf{Benchmark results.}
Table~\ref{tab:gamerealm-benchmark} reports all methods, including fine-tuned variants marked $^{*}$. Without GT geometry, MoSE3 reaches overall AJ$_{3D}$/APD$_{3D}$ of 19.4/31.4. TrackEverything+$\pi^3$ improves from 17.4/28.6 to 19.3/31.5 after fine-tuning, while OA rises from 65.5 to 72.9. Track4World achieves strong 2D localization on successfully evaluated sequences, but its incomplete coverage, particularly 2/20 Long-horizon and 14/30 Transportation sequences, must be considered when interpreting its averages. The table retains these coverage counts alongside the results.

With GT geometry, fine-tuned TAPIP3D-camera achieves the highest overall AJ$_{3D}$ and APD$_{3D}$, improving from 25.1/41.9 to 40.2/57.0. Its world-space counterpart improves from 25.8/43.2 to 39.5/55.6. Fine-tuning also raises overall OA from 59.5 to 81.9 in camera space and from 60.1 to 81.2 in world space. The gains span both Interaction and Locomotion, with particularly strong recovery on Hard and Long-horizon: camera-space AJ$_{3D}$ rises from 4.1 to 22.6 and from 8.3 to 27.2, respectively.

Among original models, Hard and Long-horizon yield the lowest APD$_{2D}$ scores for every With GT method. TrackEverything+GT illustrates the difficulty of sustained motion, with APD$_{3D}$ falling from 79.2 on Embodied to 0.4 on Long-horizon. CoTracker3 outperforms both original TAPIP3D variants on Hard and Long-horizon, highlighting the importance of temporal association and occlusion recovery. Plant and Group expose a complementary challenge: similar-looking targets overlap in the image but lie at different depths. Original TAPIP3D outperforms CoTracker3, DELTA, and SpatialTrackerV2 in 3D localization on both subsets, supporting matching neighborhoods organized by 3D distance rather than image proximity~\citep{zhang2026tapip3d}.

\noindent\textbf{Fine-tuning and transfer setup.}
Starred models are fine-tuned on our tracking data while retaining their original architectures and evaluation input conditions. The public-benchmark tables compare them with published baselines from the respective method papers~\citep{zhang2026tapip3d,jain2026trackeverything}; TAPIP3D baselines on SynthVerse follow SynthVerse~\citep{Zhao2026SynthVerse}. Synthetic evaluation uses GT geometry. Real-world evaluation uses MegaSaM~\citep{li2025megasam} geometry for TAPIP3D and $\pi^3$~\citep{wang2026pi} geometry for TrackEverything. TrackEverything is evaluated on 50 full-length minival videos per dataset. Public-benchmark AJ and OA thresholds are selected after sequence-score averaging. Avg. weights the three datasets equally, except that the synthetic TAPIP3D baseline retains its published average. $\Delta$ reports percentage-point differences between the displayed fine-tuned and baseline scores.

\begin{table*}[!t]
\centering
\caption{\textbf{Public synthetic benchmark results on SynthVerse~\citep{Zhao2026SynthVerse}, Dynamic Replica~\citep{karaev2023dynamicreplica}, and LSFOdyssey~\citep{wang2025scenetracker}.} $^{*}$: fine-tuned on our tracking data. $\Delta$: fine-tuned minus published baseline (pp). Yellow marks the higher score within each model pair; ties at the displayed precision are unshaded.}
\label{tab:synthetic_tracking_comparison}
\setlength{\tabcolsep}{3pt}
\renewcommand{\arraystretch}{1.05}
\resizebox{\textwidth}{!}{%
\begin{tabular}{l|ccccc|ccccc|ccccc|ccccc}
\toprule
\multirow{2}{*}{Methods} & \multicolumn{5}{c|}{SynthVerse-mAverage} & \multicolumn{5}{c|}{Dynamic Replica} & \multicolumn{5}{c|}{LSFOdyssey} & \multicolumn{5}{c}{Avg.} \\
 & $\mathrm{AJ}_{3D}\uparrow$ & $\mathrm{APD}_{3D}\uparrow$ & $\mathrm{AJ}_{2D}\uparrow$ & $\mathrm{APD}_{2D}\uparrow$ & $\mathrm{OA}\uparrow$ & $\mathrm{AJ}_{3D}\uparrow$ & $\mathrm{APD}_{3D}\uparrow$ & $\mathrm{AJ}_{2D}\uparrow$ & $\mathrm{APD}_{2D}\uparrow$ & $\mathrm{OA}\uparrow$ & $\mathrm{AJ}_{3D}\uparrow$ & $\mathrm{APD}_{3D}\uparrow$ & $\mathrm{AJ}_{2D}\uparrow$ & $\mathrm{APD}_{2D}\uparrow$ & $\mathrm{OA}\uparrow$ & $\mathrm{AJ}_{3D}\uparrow$ & $\mathrm{APD}_{3D}\uparrow$ & $\mathrm{AJ}_{2D}\uparrow$ & $\mathrm{APD}_{2D}\uparrow$ & $\mathrm{OA}\uparrow$ \\

\midrule
\multicolumn{21}{l}{\textit{TAPIP3D-camera}} \\
TAPIP3D  & 33.3 & 46.2 & 47.1 & \first{66.9} & 79.4 & 53.7 & 70.8 & 64.6 & 84.7 & 84.7 & 68.3 & 83.2 & 76.0 & 91.2 & 86.2 & 51.8 & 66.7 & 62.6 & 80.9 & 83.4 \\
TAPIP3D$^{*}$  & \first{36.4} & \first{48.0} & \first{48.1} & 65.9 & \first{79.9} & \first{61.5} & \first{75.4} & \first{69.8} & \first{85.4} & \first{87.8} & \first{73.5} & \first{86.0} & \first{80.2} & \first{92.4} & \first{90.4} & \first{57.2} & \first{69.8} & \first{66.0} & \first{81.2} & \first{86.1} \\
$\Delta$ & $+3.1$ & $+1.8$ & $+1.0$ & $-1.0$ & $+0.5$ & $+7.8$ & $+4.6$ & $+5.2$ & $+0.7$ & $+3.1$ & $+5.2$ & $+2.8$ & $+4.2$ & $+1.2$ & $+4.2$ & $+5.4$ & $+3.1$ & $+3.4$ & $+0.3$ & $+2.7$ \\
\midrule
\multicolumn{21}{l}{\textit{TAPIP3D-world}} \\
TAPIP3D  & 33.3 & 46.4 & 47.4 & \first{67.2} & 79.8 & 55.5 & 72.8 & 66.2 & 85.7 & 85.3 & 72.2 & 85.8 & 78.5 & 92.8 & 86.9 & 53.7 & 68.3 & 64.0 & 81.9 & 84.0 \\
TAPIP3D$^{*}$  & \first{37.6} & \first{49.4} & \first{49.1} & 66.6 & \first{80.8} & \first{62.8} & \first{75.7} & \first{71.1} & \first{86.6} & \first{87.5} & \first{76.2} & \first{88.1} & \first{82.2} & \first{93.7} & \first{90.5} & \first{58.9} & \first{71.1} & \first{67.5} & \first{82.3} & \first{86.3} \\
$\Delta$ & $+4.3$ & $+3.0$ & $+1.7$ & $-0.6$ & $+1.0$ & $+7.3$ & $+2.9$ & $+4.9$ & $+0.9$ & $+2.2$ & $+4.0$ & $+2.3$ & $+3.7$ & $+0.9$ & $+3.6$ & $+5.2$ & $+2.8$ & $+3.5$ & $+0.4$ & $+2.3$ \\
\bottomrule
\end{tabular}}
\end{table*}

\noindent\textbf{Transfer to synthetic environments.}
Table~\ref{tab:synthetic_tracking_comparison} shows higher AJ$_{3D}$ and APD$_{3D}$ on all three benchmarks in both coordinate settings. Average AJ$_{3D}$ rises from 51.8 to 57.2 in camera space and from 53.7 to 58.9 in world space. On Dynamic Replica, camera-space AJ$_{3D}$ increases from 53.7 to 61.5 and world-space AJ$_{3D}$ from 55.5 to 62.8. The corresponding LSFOdyssey scores increase from 68.3 to 73.5 and from 72.2 to 76.2. Average OA also improves, reaching 86.1 and 86.3. SynthVerse APD$_{2D}$ decreases slightly in both settings, while its 3D localization and visibility scores improve.

\begin{table*}[!t]
\centering
\caption{\textbf{Public real-world benchmark results on ADT~\citep{pan2023aria}, DriveTrack~\citep{balasingam2024drivetrack}, and PStudio~\citep{joo2015panoptic} from TAPVid-3D~\citep{koppula2024tapvid}.} $^{*}$: fine-tuned on our tracking data. TrackEverything~\citep{jain2026trackeverything} uses 50 full-length minival videos per dataset. $\Delta$: fine-tuned minus published baseline (pp). Yellow marks the higher score within each model pair; ties at the displayed precision are unshaded.}
\label{tab:tracking_comparison}
\setlength{\tabcolsep}{1.5pt}
\renewcommand{\arraystretch}{1.05}
\resizebox{0.9225\textwidth}{!}{%
\begin{tabular}{l|ccc|ccc|ccc|ccc}
\toprule
\multirow{2}{*}{Methods} & \multicolumn{3}{c|}{ADT} & \multicolumn{3}{c|}{DriveTrack} & \multicolumn{3}{c|}{PStudio} & \multicolumn{3}{c}{Avg.} \\
 & $\mathrm{AJ}_{3D}\uparrow$ & $\mathrm{APD}_{3D}\uparrow$ & $\mathrm{OA}\uparrow$ & $\mathrm{AJ}_{3D}\uparrow$ & $\mathrm{APD}_{3D}\uparrow$ & $\mathrm{OA}\uparrow$ & $\mathrm{AJ}_{3D}\uparrow$ & $\mathrm{APD}_{3D}\uparrow$ & $\mathrm{OA}\uparrow$ & $\mathrm{AJ}_{3D}\uparrow$ & $\mathrm{APD}_{3D}\uparrow$ & $\mathrm{OA}\uparrow$ \\

\midrule
\multicolumn{13}{l}{\textit{TAPIP3D-camera +MegaSaM}} \\
TAPIP3D  & 21.6 & 31.0 & 90.4 & 14.6 & \first{21.3} & 82.2 & 18.1 & 27.7 & 85.5 & 18.1 & 26.7 & 86.0 \\
TAPIP3D$^{*}$  & \first{24.0} & \first{32.6} & \first{93.6} & \first{14.8} & 20.7 & \first{84.3} & \first{19.1} & \first{28.5} & \first{88.0} & \first{19.3} & \first{27.2} & \first{88.6} \\
$\Delta$ & $+2.4$ & $+1.6$ & $+3.2$ & $+0.2$ & $-0.6$ & $+2.1$ & $+1.0$ & $+0.8$ & $+2.5$ & $+1.2$ & $+0.5$ & $+2.6$ \\
\midrule
\multicolumn{13}{l}{\textit{TAPIP3D-world +MegaSaM}} \\
TAPIP3D  & 23.5 & 32.8 & 91.2 & 14.9 & 21.8 & 82.6 & 18.1 & 27.7 & 85.5 & 18.8 & 27.4 & 86.4 \\
TAPIP3D$^{*}$  & \first{25.4} & \first{34.1} & \first{94.1} & \first{15.5} & \first{22.1} & \first{84.2} & \first{19.1} & \first{28.5} & \first{88.0} & \first{20.0} & \first{28.3} & \first{88.8} \\
$\Delta$ & $+1.9$ & $+1.3$ & $+2.9$ & $+0.6$ & $+0.3$ & $+1.6$ & $+1.0$ & $+0.8$ & $+2.5$ & $+1.2$ & $+0.9$ & $+2.4$ \\
\midrule
\multicolumn{13}{l}{\textit{TrackEverything +$\pi^3$}} \\
TrackEverything  & 21.1 & 31.0 & 87.9 & 17.3 & 24.7 & 84.2 & 11.4 & 18.9 & \first{86.0} & 16.6 & 24.9 & 86.0 \\
TrackEverything$^{*}$  & \first{24.9} & \first{32.8} & \first{91.0} & \first{18.8} & \first{24.9} & \first{89.3} & \first{12.9} & \first{20.5} & 85.4 & \first{18.9} & \first{26.1} & \first{88.6} \\
$\Delta$ & $+3.8$ & $+1.8$ & $+3.1$ & $+1.5$ & $+0.2$ & $+5.1$ & $+1.5$ & $+1.6$ & $-0.6$ & $+2.3$ & $+1.2$ & $+2.6$ \\
\bottomrule
\end{tabular}}
\end{table*}

\noindent\textbf{Transfer to real-world video.}
Table~\ref{tab:tracking_comparison} shows gains for both trackers. TAPIP3D's average AJ$_{3D}$ increases from 18.1 to 19.3 in camera space and from 18.8 to 20.0 in world space. All three world-space metrics improve on every dataset. In camera space, DriveTrack APD$_{3D}$ decreases from 21.3 to 20.7 despite improvements in AJ$_{3D}$ and OA. TrackEverything improves AJ$_{3D}$ and APD$_{3D}$ on all three minival subsets, increasing their averages from 16.6/24.9 to 18.9/26.1. Its average OA rises from 86.0 to 88.6, with a small decrease on PStudio from 86.0 to 85.4. These results demonstrate transfer across trackers and real-world settings while separating localization and visibility behavior.

\begin{figure*}[t]
    \centering
    \begingroup
    \normalfont\normalsize
    \newcommand{\TrackingRow}[3]{%
      \hbox to \linewidth{%
        \makebox[.14\linewidth][l]{\raisebox{-.5\height}{#1}}%
        \raisebox{-.5\height}{\includegraphics[trim={47.1273bp #2bp 0bp #3bp},clip,width=.86\linewidth]{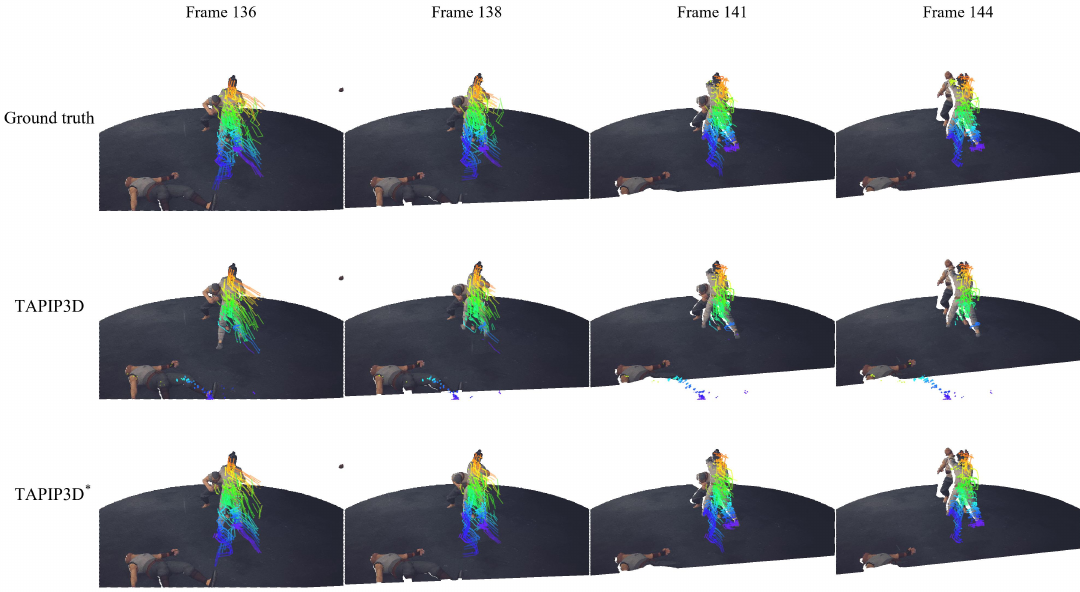}}}%
    }
    \vbox{\offinterlineskip
      \TrackingRow{Ground-truth}{182.6182}{11.1273}%
      \TrackingRow{TAPIP3D}{92.2909}{101.4545}%
      \TrackingRow{TAPIP3D\textsuperscript{*}}{1.9636}{191.7818}%
    }
    \endgroup
    \caption{3D tracking during character interaction. Top to bottom: ground truth, TAPIP3D, and TAPIP3D$^{*}$ fine-tuned on our game data. Matching colors identify the same query points across rows and frames.}
    \label{fig:sifu_tracking_comparison}
\end{figure*}

\noindent\textbf{Qualitative comparison.}
Figure~\ref{fig:sifu_tracking_comparison} complements the quantitative results with world-space trajectories from a separate game-fine-tuned checkpoint. The recording group of the displayed sequence is excluded from fine-tuning. Sixteen-frame trajectory segments are shown at four time points using shared query identities, point colors, reference visibility masks, and a fixed viewpoint, without trajectory smoothing. During character interaction, the original TAPIP3D model shifts several lower-body tracks toward another character in the foreground. TAPIP3D$^{*}$ keeps these tracks closer to the target character and the ground-truth trajectories throughout the displayed frames, preserving point association through the interaction.

\FloatBarrier

\section{Conclusion}
In this work, we introduced GameRealm, a large-scale game-derived dataset that grounds visual observations in ground-truth geometric and interactive information. By combining human-operated interaction capture with agentic automated exploration, GameRealm provides diverse and synchronized multimodal data across multiple viewing configurations. We further establish a scalable acquisition and quality assessment pipeline to ensure data integrity and cross-view consistency. Experiments across multiple representative visual learning tasks demonstrate its utility for interactive world modeling and visual geometry learning, with promising results in action-conditioned generation and cross-domain geometric generalization. In future work, we aim to enrich GameRealm with finer-grained ground-truth annotations, including human and object skeletons, pixel-level segmentation masks, and instance-level information, to support a broader range of visual learning tasks.

\bibliography{refs}

\appendix

\end{document}